\documentclass[lettersize,journal]{IEEEtran}

\usepackage{amsmath, amsfonts, bm}

\usepackage{graphicx}
\usepackage[caption=false,font=normalsize,labelfont=sf,textfont=sf]{subfig}

\usepackage{array}
\usepackage{booktabs}
\usepackage{threeparttable}
\usepackage{multirow}
\usepackage[table]{xcolor}

\usepackage[nocompress]{cite}

\usepackage[ruled,linesnumbered]{algorithm2e}

\usepackage{enumitem}
\usepackage{textcomp}
\usepackage{url}
\usepackage{verbatim}
\usepackage{bbm}
\usepackage{dsfont}
\usepackage{color}

\usepackage[
    pagebackref=true,
    breaklinks=true,
    colorlinks=true,
    bookmarks=false,
    linkcolor=red,
    anchorcolor=black,
    citecolor=green,
    urlcolor=black
]{hyperref}

\begin{document}

\title{DAPM: UAV Monocular Depth Estimation from Any Height, Pitch, Roll and FOV}

\author{
Tong~Ling, Wenhui~Diao, Member, IEEE, Yingchao~Feng, Member, IEEE, Hanbo~Bi, Zhongyan~Hou,\\ Xian~Sun, Senior Member, IEEE

\IEEEcompsocitemizethanks{

\IEEEcompsocthanksitem This work was supported by the National Natural Science Foundation of China under Grant 62425115. (Corresponding authors: W. Diao.)
\IEEEcompsocthanksitem T. Ling, W. Diao, H. Bi, Z. Hou, and X. Sun are with the Aerospace Information Research Institute, Chinese Academy of Sciences, Beijing 100190, China, also with the School of Electronic, Electrical and Communication Engineering, University of Chinese Academy of Sciences, Beijing 100190, China, also with the University of Chinese Academy of Sciences, Beijing 100190, China, and also with the Key Laboratory of Target Cognition and Application Technology (TCAT), Aerospace Information Research Institute, Chinese Academy of Sciences, Beijing 100190, China (e-mail: lingtong23@mails.ucas.ac.cn, diaowh@aircas.ac.cn, sunxian@aircas.ac.cn).
\IEEEcompsocthanksitem Y. Feng is with the Aerospace Information Research Institute, Chinese Academy of Sciences, Beijing 100190, China, and also with the Key Laboratory of Target Cognition and Application Technology (TCAT), Aerospace Information Research Institute, Chinese Academy of Sciences, Beijing 100190, China.

}
}

\markboth{Journal of \LaTeX\ Class Files,~Vol.~14, No.~8, August~2021}%
{Shell \MakeLowercase{\textit{et al.}}: A Sample Article Using IEEEtran.cls for IEEE Journals}


\maketitle

\begin{abstract}

Monocular depth estimation is a fundamental prerequisite for 3D reconstruction and autonomous navigation in Unmanned Aerial Vehicles (UAVs). In practical deployments, UAVs operate under highly dynamic camera poses characterized by continuous variations in height, pitch, roll, and field of view (FOV). Existing monocular depth estimation methods frequently fail to generalize across such diverse perspectives and the expansive scale of depth distributions inherent in aerial scenes. To address these challenges, we establish a quantitative representation of UAV viewing angles through rigorous theoretical analysis, deriving the geometric correspondence between viewing angles and view distances using the ground plane as a reference for observation. Building upon this, we propose Depth Estimation for Any Perspectives Model (DAPM), representing the first monocular framework specifically designed for UAV aerial imagery to jointly estimate camera pose and depth under continuously varying viewpoints. Specifically, we introduce an Ideal Ground Depth (IGD) module that leverages the derived geometric relationships between UAV perspectives and view distances to implement dense camera-pose supervision and enhance depth features. And we further develop a coarse-to-fine Progressive Quantization Bins (PQB) module. By incorporating progressive supervision and hierarchical quantization bins, the PQB module enables robust estimation in complex UAV aerial imagery. To evaluate the proposed framework, we present the UAV Any Perspectives Depth (UAPD) dataset, featuring comprehensive and continuous distributions of pose parameters. Experimental results on UAPD demonstrate that DAPM achieves state-of-the-art performance across both depth and camera-pose estimation metrics. The source code and datasets are available at: https://github.com/ThisIsLT/DAPM.

\end{abstract}


\section{Introduction}
\IEEEPARstart{W}{ith} the rapid advancement of unmanned aerial vehicle (UAV) technology, the low-altitude economy has become a focal point in contemporary societal development. Accordingly, research on low-altitude remote sensing imagery has seen significant growth in recent years \cite{wang2024drones, tian2024ucdnet, diao2024ringmo, he2025progressive, hu2025ringmo, ren2025supermot}. 
Depth estimation from UAV imagery has emerged as a key enabler for applications such as autonomous navigation \cite{xiao2025uav, javaid2025explainable, farid2025improved}, 3D reconstruction \cite{li2024single, maboudi2023review, panagiotopoulou2023super}, and embodied intelligence \cite{yao2024aeroverse, sapkota2025uavs, guo2026bedi}.

\begin{figure}
\centering
\includegraphics[width=0.9\linewidth]{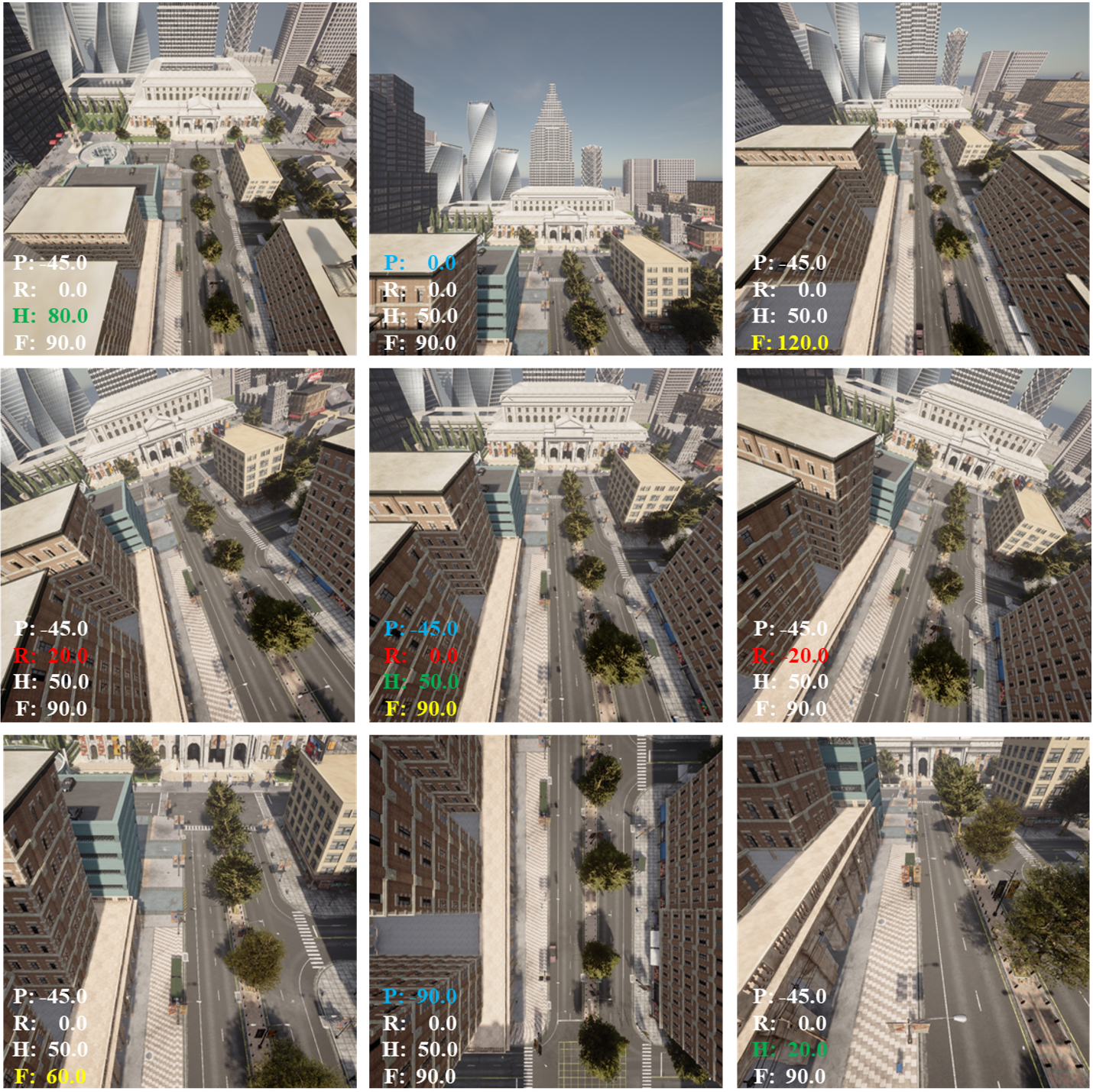}
\caption{The monocular UAV images exhibit distinct visual characteristics under varying \textbf{Pitch}, \textbf{Roll}, \textbf{Height}, and \textbf{FoV} conditions. The figure presents a comparative visualization of UAV imagery under each of these perspective variations.}
\label{fig:ds}
\end{figure}

Unlike autonomous driving or indoor scenarios where the imaging viewpoints are relatively constrained, as illustrated in Fig.~\ref{fig:ds}, UAV imagery involves a wide and continuous distribution of camera poses, including height, pitch, roll, and field of view (FOV). This results in highly dynamic scene geometry and depth distribution, posing considerable challenges for accurate monocular depth estimation across arbitrary UAV viewpoints. While recent advancements have introduced diffusion-based depth estimation frameworks \cite{song2025depthmaster, gui2025depthfm}, their prohibitive computational overhead and substantial parameter counts result in slow inference latencies, rendering them impractical for real-time deployment on resource-constrained UAV platforms. Other methods \cite{bhat2021adabins, li2024binsformer} mainly focus on reformulating continuous depth prediction as a classification–regression task: the model first predicts a probability distribution over discrete depth intervals (bins) for each pixel, and then computes the final depth value as a linear combination of the bin centers weighted by these probabilities. However, when applied to aerial imagery, existing methods struggle to handle the diverse camera viewpoint distributions and the large-scale depth ranges inherent in aerial scenes, resulting in insufficient generalization performance.

The key to addressing these issues lies in establishing a quantitative correlation between depth information and UAV viewpoint parameters, thereby enhancing the model’s capability by improving its understanding of aerial viewpoints and viewing distances. For instance, roll characterizes the rotational properties of the horizon in aerial imagery, where its variations induce corresponding tilts in the depth distribution. Grounded in this observation, a straightforward approach is to integrate depth estimation with pose estimation, enabling a synergistic enhancement through their intrinsic coupling. 
Consequently, this work focuses on addressing two pivotal problems: (1) how to quantitatively characterize the relationship between depth information and UAV viewpoint parameters; and (2) how to effectively leverage such information to augment the model's estimation capabilities. 

To address Problem (1), this paper theoretically demonstrates that a coupling relationship exists between depth and camera pose in aerial images. First, through an analysis of the camera intrinsic and extrinsic parameter matrices, parameters unrelated to aerial operations, such as those representing specific camera hardware characteristics, are excluded. We show that the aerial imaging viewpoint can be effectively represented using only four parameters: $[\text{FoV}, \text{Pitch}, \text{Roll}, \text{Height}]$. Based on these parameters, we further derive a quantitative geometric relationship between the aerial viewpoint and the depth–distance distribution when using an ideal ground plane as the observation reference. To address Problem (2), we observe that the ground-plane depth map provides a useful geometric prior for depth estimation while simultaneously encoding imaging geometry and viewpoint information. Therefore, introducing ideal ground depth as an additional feature input can improve depth estimation in complex scenes. However, although onboard sensors (GPS/IMU) provide pose data, their reliability degrades under adverse conditions, making image-based pose estimation essential for improving the model's robustness and geometric understanding. Meanwhile, existing pose estimation methods \cite{jin2023perspective, veicht2024geocalib} are primarily designed for ground or indoor scenes and fail to generalize to the large-scale geometric variations present in UAV imagery. Therefore, this paper incorporates camera pose estimation capabilities and also represents the first work to investigate camera pose estimation specifically for aerial imagery.

\begin{figure}
\centering
\includegraphics[width=0.9\linewidth]{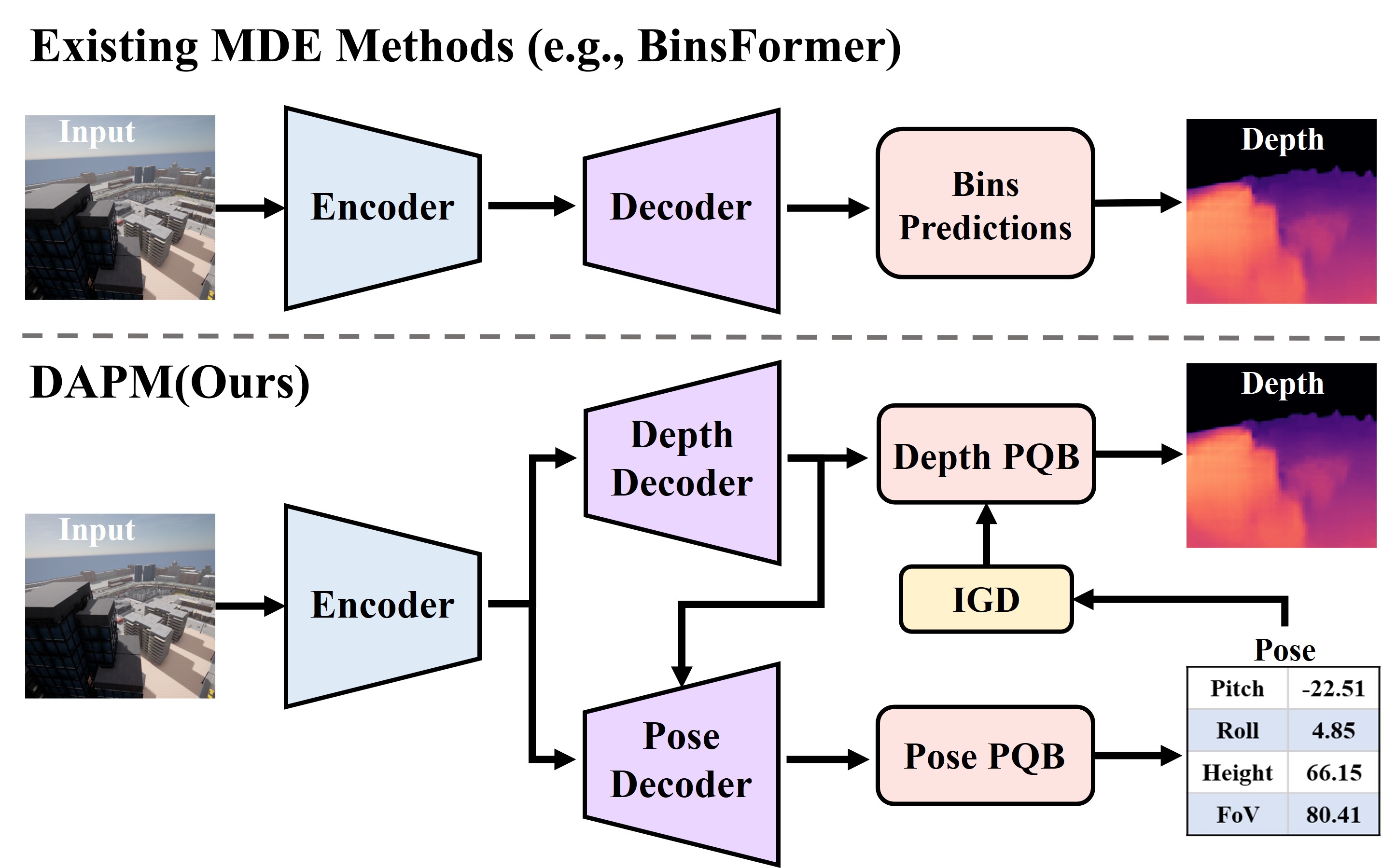}
\caption{Compared with BinsFormer, DAPM employs a coarse-to-fine pixel-level binning strategy, significantly improving depth estimation accuracy for aerial images with large scale variations. It also integrates camera pose estimation with pose-aligned ground features, leveraging the geometric relationship between depth and pose to further enhance precision.}
\label{fig:dvsb}
\end{figure}

Based on this theoretical analysis, we adopt a coupled depth-and-pose estimation strategy, where the estimated camera pose information is leveraged to improve depth estimation performance for aerial imagery with complex viewpoint variations. Consequently, as illustrated in Fig.~\ref{fig:dvsb}, taking BinsFormer as a representative baseline, our method simultaneously estimates camera pose information and utilizes these parameters to enhance depth estimation.

To address these challenges, we propose the Depth Estimation for Any Perspectives Model (DAPM), a unified framework specifically designed for UAV imagery. In DAPM, depth estimation serves as the primary task, while camera pose acts as auxiliary supervision. By exploiting the geometric interdependence between the two, the model achieves enhanced geometric structural representation learning, thereby improving its robustness and generalization across diverse pose conditions. First, to capture the theoretical relationship between UAV viewing angle and view distance, we derive the depth-distance distribution of the ground plane under arbitrary UAV viewing angles, using the ground plane as the observational reference. Building upon this, we propose an Ideal Ground Depth (IGD) module. This module derives the ideal ground depth using the estimated pose parameters and incorporates it both as a pose-guided dense supervision signal and as input for depth feature refinement, thereby enhancing the model's capability by introducing sky-ground background prior features and viewing perspective representations. To address the dynamic variation of depth distributions caused by changing camera viewpoints, we introduce a Progressive Quantization Bins (PQB) strategy that utilizes pixel-wise bins rather than globally shared bins to avoid mutual interference between pixels with different depth scales. This strategy gradually increases the resolution of discrete depth and pose bins, providing a coarse-to-fine supervision signal that guides the network toward more precise predictions.

Current monocular depth estimation methods are predominantly trained and evaluated on datasets with limited viewpoint variability, particularly those with restricted height changes \cite{geiger2013vision, silberman2012indoor, feng2023height, miclea2021monocular}. This limitation makes them difficult to generalize to real-world UAV scenarios, characterized by continuous and wide-ranging camera pose variations. Even among existing UAV datasets that incorporate viewpoint changes, significant limitations remain: they either lack annotated camera parameters \cite{liu2020novel} or provide only limited types of pose variation \cite{rizzoli2023syndrone}, which restricts the investigation into how pose changes affect depth estimation. To comprehensively evaluate performance under arbitrary UAV viewpoints, we construct a new dataset named UAPD, generated using the CARLA simulation platform. UAPD contains 42k images with continuously distributed height, pitch, roll, and field-of-view parameters, along with ground-truth depth maps and pose annotations, enabling effective evaluation of UAV depth estimation and camera pose estimation tasks.

To validate the effectiveness of the proposed DAPM, we first conduct comprehensive experiments on the UAPD dataset. The results show that DAPM achieves state-of-the-art performance on both depth and pose estimation, significantly outperforming existing mainstream approaches and confirming the strength of the proposed framework. Overall, DAPM is the first monocular UAV imagery framework that jointly tackles continuous viewpoint-aware depth estimation and camera pose estimation in a unified architecture. In summary, the contributions of this paper can be summarized as follows:

\begin{itemize}

\item \textbf{A Unified Joint Framework:} We propose DAPM, the first framework for joint depth and camera pose estimation from UAV imagery under arbitrary viewpoints, which leverages the geometric coupling between pose and depth distributions to mutually enhance both tasks.

\item \textbf{Theoretical Geometric Coupling:} We derive the geometric relationship between viewing angle and depth-distance distribution from the ground plane, and propose an Ideal Ground Depth module that integrates sky–ground priors and viewpoint representations to enhance depth estimation and enable dense 2D pose supervision.

\item \textbf{Progressive Quantization Bins (PQB):} We propose a coarse-to-fine Progressive Quantization Bins strategy that incrementally increases bin resolution to effectively handle large-scale variations and progressively refine camera pose and depth estimation accuracy.

\item \textbf{The UAPD Dataset:} We propose the UAPD dataset, comprising 42k aerial images with evenly distributed viewpoints, along with depth ground truth and camera parameters, for evaluating depth and camera pose estimation on drone imagery with continuous perspective variations.

\end{itemize}

\section{Related Work}

\subsection{Monocular Depth Estimation}

Monocular depth estimation has evolved from multi-scale CNNs \cite{eigen2014depth, wang2020cliffnet} to Transformer-based \cite{ranftl2021vision} and hybrid classification-regression architectures \cite{johnston2020self, bhat2021adabins, li2024binsformer}. While recent foundation models \cite{yang2024depth2, piccinelli2025unidepthv2, wang2025vggt} demonstrate strong capabilities, their massive parameter counts differ from the lightweight requirements of UAV platforms. Furthermore, standard datasets \cite{geiger2013vision, silberman2012indoor} lack the substantial viewpoint variations inherent to aerial imagery. Although some aerial studies \cite{feng2023height, miclea2021monocular, li2023hierarchical} and UAV-specific methods \cite{madhuanand2021self, shimada2022pix2pix, yu2023scene} exist, they generally do not investigate the theoretical relationship between depth prediction and UAV viewpoint changes.


\subsection{Camera Pose Estimation}


Deep learning has significantly advanced camera pose estimation. Key methodologies range from direct 6-DoF regression like PoseNet \cite{kendall2015posenet} to high-precision scene coordinate regression such as DSAC \cite{brachmann2021visual}, which utilizes RANSAC optimization. To enhance generalization, hierarchical structures like HSC-Net \cite{li2020hierarchical} and panoramic knowledge distillation have been developed. Recent innovations include Perspective Fields \cite{jin2023perspective}, which models local geometry via pixel-level vectors, and GeoCalib \cite{veicht2024geocalib}, which leverages geometric constraints for self-supervised optimization. Despite these gains, current models primarily target indoor or street-level views, remaining inadequate for the wide fields of view and long-range perspectives typical of UAV imagery.

\subsection{UAV Depth Estimation datasets}

For the task of depth estimation from UAV imagery, several datasets have been proposed. The Mid-Air dataset \cite{fonder2019mid} uses the Unreal Engine with the AirSim plugin to simulate flight trajectories under varying climatic and seasonal conditions, and supports multitask learning objectives such as SLAM and visual localization. WildUAV \cite{florea2021wilduav} provides real-world RGB images captured in natural scenes along with high-quality depth ground truth generated via photogrammetry, making it well-suited for assessing the generalization of depth‐estimation methods over complex terrain. UEMM-Air \cite{yao2024uemm} generates high-precision synthetic images via the CARLA and Unreal platforms, respectively. DDOS \cite{kolbeinsson2024ddos}, which provides photorealistic UAV imagery with depth and obstacle segmentation labels, specifically targeting thin structures. However, each dataset remains confined to one or a few fixed camera tilt angles and lacks continuous distributions of field-of-view and flight height, limiting their utility for studying depth estimation under multi-viewpoint variation. In summary, existing UAV datasets suffer from limitations: either lacking annotated camera parameters \cite{liu2020novel} or providing only limited types of pose variation \cite{rizzoli2023syndrone}, which restricts the investigation into how pose changes affect depth estimation.


\subsection{Depth and Camera Pose Joint Estimation}

Depth and camera pose joint estimation has emerged as a crucial research direction in computer vision and robotics. UnDEMoN \cite{babu2018undemon} pioneered this approach, which was subsequently improved for dynamic scenes by incorporating optical flow \cite{wang2020unsupervised} and geometric consistency constraints \cite{zhang2023unsupervised}. Others extended the framework with ground normal estimation \cite{zhang2020joint}. In challenging endoscopic scenarios, methods addressed texture sparsity and illumination via appearance flow \cite{shao2021self} and self-attention mechanisms \cite{li2025enhanced}. Geometric approaches have also explored relative pose estimation using non-metric depth \cite{eichhardt2020relative} or joint scale-shift optimization \cite{ding2025fixing}. In the domain of UAV imagery, Zhang et al. \cite{zhang2022uav} utilized weighted losses for stabilization, while Chang et al. \cite{chang2023mixed} focused on multi-scale feature enhancement. Other studies integrated sensor data to recover metric scale \cite{shimada2023fast, florea2025tandepth}. Notably, end-to-end supervised depth and pose joint methods remain largely unexplored in aerial imagery research, with most approaches relying on self-supervised or semi-supervised training. Therefore, we propose DAPM, the first end-to-end supervised framework for joint depth and pose estimation specifically tailored to UAV imagery.

\section{Methodology}

\subsection{Methodology Overall}

The methodology of this paper is organized as follows. To address the multi-viewpoint and long-scale characteristics of aerial imagery, we first derive the relationship between UAV viewpoint variations and viewing-distance distributions. Based on this analysis, we propose the DAPM model, which integrates an Ideal Ground Depth module derived from this geometric relationship to jointly improve depth and pose estimation, along with a Progressive Quantization Bins module that enhances accuracy through hierarchical supervision. We further introduce UAPD, the first UAV dataset with continuously distributed viewpoints, for evaluating depth and pose estimation. Overall, our method comprises three main components:

\begin{itemize}

\item \textbf{Theoretical Relation Between UAV Perspective and Viewing-Distance Distribution}:
By analyzing the camera parameters, we show that UAV perspectives are fully determined by Height, FoV, Pitch, and Roll, and derive their relationship with viewing-distance distributions using the ground plane as reference.

\item \textbf{Depth Estimation for Any-Perspective Model}:
We present the overall model architecture and detail the designs of the Ideal Ground Depth and Progressive Quantization Bins modules, together with the loss function for joint depth and pose estimation.

\item \textbf{UAV Any-Perspective Depth Dataset}:
We introduce the UAPD dataset and its construction pipeline, and demonstrate through comparison with existing datasets that it offers the most comprehensive coverage of UAV viewpoint variations.
\end{itemize}

\subsection{Theoretical Relation Between UAV Perspective and Viewing-Distance Distribution}

\subsubsection{Key Parameters Influencing Viewing Angles}

The viewpoint of a UAV image is fundamentally determined by the camera pose. The camera parameters are primarily represented by the intrinsic matrix $K$ and the extrinsic matrix $T$\cite{kendall2015posenet, veicht2024geocalib}. We can construct the parameter matrix $P$, jointly represented by $K$ and $T$, using a total of 10 parameters $\{{FoV}_x, {FoV}_y, c_x, c_y, Pitch, Roll, Yaw, t_x, t_y, t_z\}$, as is provided in Eq.~\eqref{eq:1}:

\begin{equation}\label{eq:1}
\begin{gathered}
P = 
\begin{bmatrix}
K & T
\end{bmatrix}
=
\underbrace{
\begin{bmatrix}
f_x & 0 & c_x \\
0 & f_y & c_y \\
0 & 0 & 1
\end{bmatrix}
}_{{FoV}_x, {FoV}_y, c_x, c_y}
\underbrace{
\begin{bmatrix}
r_{11} & r_{12} & r_{13} \\
r_{21} & r_{22} & r_{23} \\
r_{31} & r_{32} & r_{33}
\end{bmatrix}
}_{R = R_{yaw}R_{pitch}R_{roll}}
\begin{bmatrix}
t_x \\
t_y \\
t_z
\end{bmatrix}
\\
\end{gathered}
\end{equation}


Here, the intrinsic matrix $K$ represents the internal mapping of the camera, where $f_x$ and $f_y$ denote the focal lengths along the $x$ and $y$ axes, which can be directly derived from $\text{FoV}_x$ and $\text{FoV}_y$, and $(c_x, c_y)$ is the principal point. The extrinsic matrix $T$ describes the rigid-body transformation from the world coordinate system to the camera system, where $R$ is a $3 \times 3$ rotation matrix obtained by multiplying the three rotation matrices $R_{yaw}R_{pitch}R_{roll}$ defining the camera orientation, and $\mathbf{t} = [t_x, t_y, t_z]^T$ is the translation vector representing the camera position in world coordinates. Given that the primary objective of this study is to investigate the impact of viewpoint shifts on visual distance during UAV operations, the inherent attributes of the specific camera hardware are of secondary concern. Through theoretical analysis, we demonstrate that the parameters truly reflecting the characteristics of aerial remote sensing can be reduced via simplification and transformation from 10 parameters to 4 parameters: $\{\text{FoV}, \text{Pitch}, \text{Roll}, \text{Height}\}$. We present the derivation processes for the intrinsic matrix $K$, the translation vector $\mathbf{t}$, and the rotation matrix $R$ respectively:\\



\noindent \textbf{a. Influence of ${FoV}_x$, ${FoV}_y$, $c_x$ and $c_y$ in $K$:\\}
The intrinsic matrix $K$ encodes the hardware characteristics. To eliminate the influence of non-aerial operations, we assume a unit aspect ratio, implying that $f_x = f_y$. Consequently, the intrinsic matrix is simplified to:

\begin{equation}\label{eq:2}
K = 
\begin{bmatrix}
f & 0 & c_x \\
0 & f & c_y \\
0 & 0 & 1
\end{bmatrix}
\end{equation}

Similarly, assuming that $c_x$ and $c_y$ are aligned with the center of the camera's optical axis, we can define the origin of the image coordinate system at the optical center, yielding:

\begin{equation}\label{eq:3}
K = 
\begin{bmatrix}
f & 0 & 0 \\
0 & f & 0 \\
0 & 0 & 1
\end{bmatrix}
\end{equation}

This results in a simplified intrinsic matrix under ideal camera conditions. In other words, with the exception of zooming operations, the remaining parameters in the intrinsic matrix characterize the physical properties of the camera equipment itself, rather than the unique geometric characteristics of aerial remote sensing. Furthermore, the relationship between the focal length $f$ and the Field of View ($\text{FoV}$) is defined as follows:


\begin{equation}\label{eq:4} 
\text{FoV} = \underbrace
{2 \arctan \left( \frac{D}{2f} \right)
}_{\text{as } D, f \neq 0, \text{ FoV } > 0} 
\end{equation}

where $D$ represents the pixel dimension of the image. Compared to the focal length $f$, the $\text{FoV}$ provides a more intuitive representation of the visible spatial extent. Since there is a direct mathematical mapping between these two parameters, we employ $\text{FoV}$ to characterize the camera's zooming operations.\\

\noindent \textbf{b. Influence of $t_x$, $t_y$ and $t_z$ in $\mathbf{t}$:\\}
Then we simplify the extrinsic matrix, which defines the rigid-body transformation. Since moving the UAV lens forward, backward, left, or right does not alter the perspective geometry or the scale relationships of objects, whereas the vertical coordinate $t_z$ determines the distance to the ground and dictates how flat objects appear. Therefore, as illustrated in Fig.~\ref{fig:ll2}, we establish a specific world coordinate system where the origin ($O$) is anchored on the ground directly beneath the UAV (the nadir point), and thus $t_x$ and $t_y$ are always zero at the origin. In this frame, the $Z$-axis represents altitude, and the $X$-axis is aligned with the horizontal projection of the camera's optical axis. Under this definition, the translation vector $\mathbf{t}$ is strictly determined by the flight height $h$:

\begin{equation}\label{eq:5}
\mathbf{t} = \begin{bmatrix} t_x \\ t_y \\ t_z \end{bmatrix} = \begin{bmatrix} 0 \\ 0 \\ h \end{bmatrix}.
\end{equation}\\

Consequently, for monocular aerial imagery, when analyzing only the impact of the shooting viewpoint on image content, the translation vector $\mathbf{t}$ can be fully characterized by the height $h$ alone.\\

\noindent \textbf{c. Influence of Pitch($\varphi$), Roll($\phi$) and Yaw($\psi$) in $R$:\\}
Since rotating the UAV camera to the left or right, similar to translating the camera forward, backward, left, or right, does not alter the perspective geometry or the scale relationships of objects, yaw ($\psi$) does not affect the imaging characteristics of aerial imagery. Furthermore, by aligning the coordinate system's $X$-axis with the optical axis projection, the relative yaw is effectively normalized to zero ($\psi = 0$). Consequently, the rotation matrix $R$ is determined solely by the Pitch ($\varphi$) and Roll ($\phi$). Since $\psi = 0$, the rotation matrix around the $Z$-axis, $R_{yaw}(\psi)$, simplifies as follows:

\begin{equation}\label{eq:6}
R_{yaw}(\psi) = 
\begin{bmatrix}
\cos(\psi) & -\sin(\psi) & 0 \\
\sin(\psi) & \cos(\psi) & 0 \\
0 & 0 & 1
\end{bmatrix}
= 
\begin{bmatrix}
1 & 0 & 0 \\
0 & 1 & 0 \\
0 & 0 & 1
\end{bmatrix}.
\end{equation}

As $R_{yaw}(\psi)$ reduces to the identity matrix, the calculation of the total rotation matrix $R$ is simplified to:

\begin{equation}\label{eq:7}
R = R_{roll}(\phi) R_{pitch}(\varphi) R_{yaw}(\psi) = R_{roll}(\phi) R_{pitch}(\varphi).
\end{equation}



Thus, for monocular aerial imagery, since variations in Yaw ($\psi$) do not affect the geometric imaging of the scene, the rotation matrix $R$ can be derived using only Pitch ($\varphi$) and Roll ($\phi$).




As indicated by Eq.~\eqref{eq:7}, the extrinsic matrix $T$ can be represented solely by Pitch ($\varphi$), Roll ($\phi$), and Height ($h$), while Eq.~\eqref{eq:4} demonstrates that the intrinsic matrix $K$ can be characterized by the FoV. Through this reformulation, the intrinsic and extrinsic parameter matrices of the UAV camera, representing a quantitative characterization of the aerial perspective, can be derived \textbf{solely from the parameter $[\text{FoV}, \text{Pitch}, \text{Roll}, \text{Height}]$}. While aerial observation geometry can theoretically be modeled via standard camera parameter matrix transformations, such a representation suffers from parameter redundancy and lacks intuitive interpretability. Crucially, the matrix approach fails to decouple the influence of individual parameter variations on the observation viewpoint. Therefore, to analyze how each parameter within $[\text{FoV}, \text{Pitch}, \text{Roll}, \text{Height}]$ independently affects the aerial observation geometry, we adopt the ground plane as the canonical observation scene. By modeling the visual distance relationship between the UAV camera and the ground, we derive a quantitative and geometrically interpretable representation of the aerial viewpoint.\\

\subsubsection{Angle-Distance Derivation in Ground Observation}

\label{subsec:addgo}

\begin{figure}[t]
\centering
\includegraphics[width=0.8\linewidth]{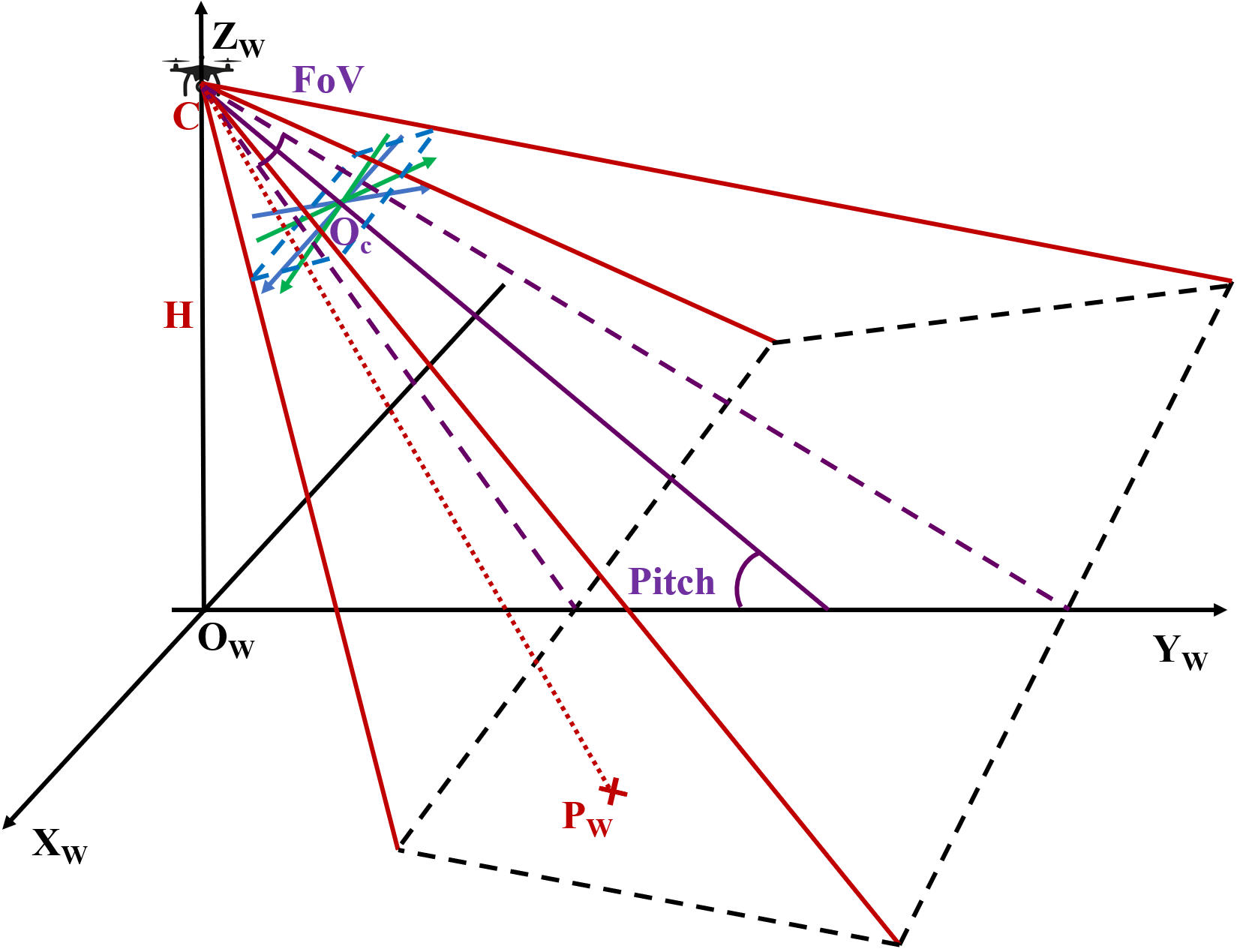}
\caption{Illustration of the ground observation range of a UAV in the world coordinate system. The ground plane position $P_W$ corresponding to any pixel $P_C$ in the world coordinate system can be obtained, and the distance from $P_W$ to the camera $C$ can be derived. The origin of both the camera and image coordinate systems is denoted by $O_C$, with further complexities detailed in Fig.~\ref{fig:ll3}.}
\label{fig:ll2}
\end{figure}


We posit zero lens distortion, a principal point aligned with the image center, and identical horizontal and vertical image dimensions $D$ (in pixels). Under these constraints, we focus exclusively on analyzing the angular characteristics and depth distribution of the UAV imagery, omitting device-specific distortion calibration parameters.

\begin{figure}[t]
\centering
\includegraphics[width=0.8\linewidth]{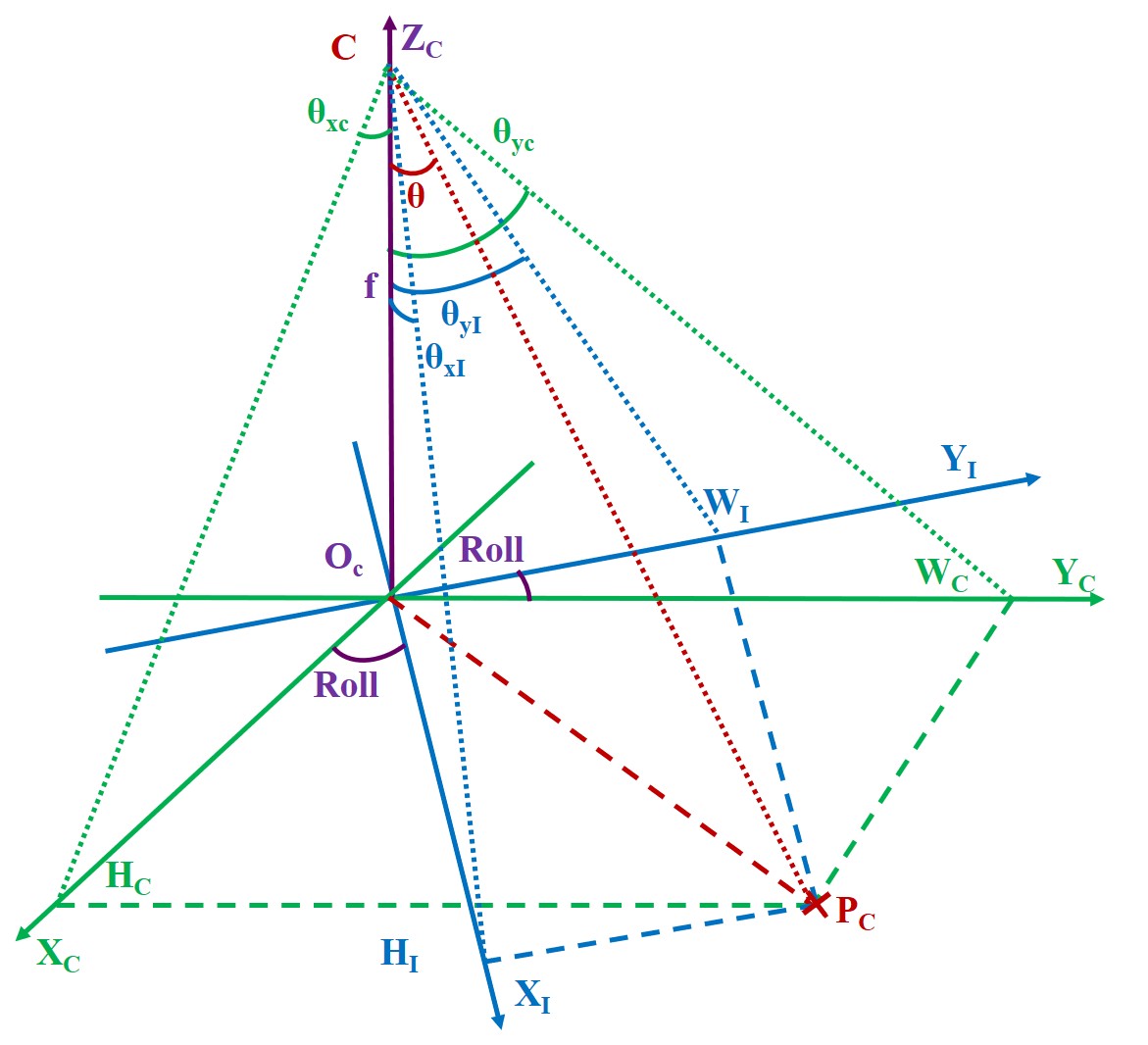}
\caption{For any pixel $P_C$ with vertical and horizontal offsets denoted as $H_C$ and $W_C$, respectively, we apply a roll rotation to transform these into the image coordinate system, yielding the coordinates $(H_{I}, W_{I})$. Then, utilizing the focal length $f$, we determine the angular deviations $\theta_{x_I}$ and $\theta_{y_I}$ between the ground position corresponding to the pixel and the camera's optical axis.}
\label{fig:ll3}
\end{figure}

\begin{figure}
\centering
\includegraphics[width=0.95\linewidth]{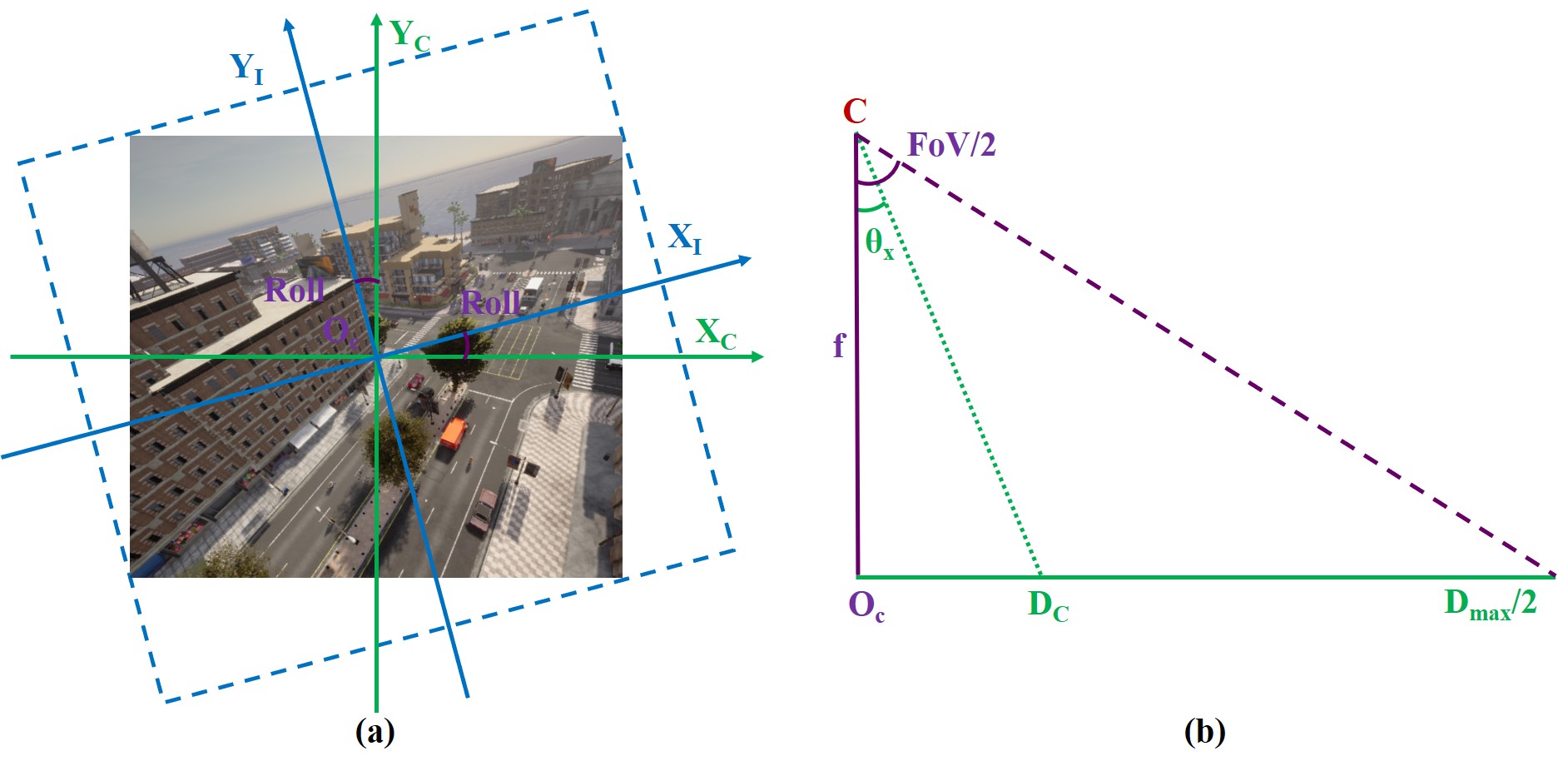}
\caption{(a) Relationship between pixel angle to optical axis and fov. (b) The camera coordinate system and the image coordinate system are transformed through a roll rotation transformation.}
\label{fig:ll1}
\end{figure}

As illustrated in Fig.~\ref{fig:ll2}, to obtain the distance from the lens to the ground corresponding to a specific pixel, we must first determine the intersection between the ground plane and the line of sight defined by the lens center and the pixel. We characterize this line by its angular deviation relative to the optical axis and project this angle onto the world coordinate system axes. This yields the ground coordinates and, subsequently, the depth distance.

Centering the camera coordinate system at the principal point, consider an arbitrary pixel $P_C$ with vertical and horizontal offsets $H_C$ and $W_C$ (in pixels). As shown in Fig.~\ref{fig:ll3} and Fig.~\ref{fig:ll1}(a), to align with the world coordinate system, we first apply a roll rotation $\phi$. The rotation matrix $R_{\mathrm{roll}}$ is defined as:
\begin{equation}\label{eq:10}
R_{\mathrm{roll}} = \begin{bmatrix}
\cos \phi & -\sin \phi \\
\sin \phi & \cos \phi
\end{bmatrix}
\end{equation}

Applying this transformation yields the rotated image coordinates $(H_{I}, W_{I})$:
\begin{equation}\label{eq:11}
\begin{bmatrix}
H_{I} \\[2pt]
W_{I}
\end{bmatrix}
=
R_{\mathrm{roll}}
\begin{bmatrix}
H_{C} \\[2pt]
W_{C}
\end{bmatrix}
\end{equation}

This demonstrates that the \textbf{Roll($\phi$) characterizes the rotation of the skyline; a non-zero roll will result in a slanted depth distribution.} In this rotated system, the tangential components of the deviation angles are:
\begin{equation}\label{eq:12}
\left\{
\begin{aligned}
\tan \theta_{x_I} &= \frac{H_I}{f}, \\
\tan \theta_{y_I} &= \frac{W_I}{f}.
\end{aligned}
\right.
\end{equation}

\begin{figure}
\centering
\includegraphics[width=1\linewidth]{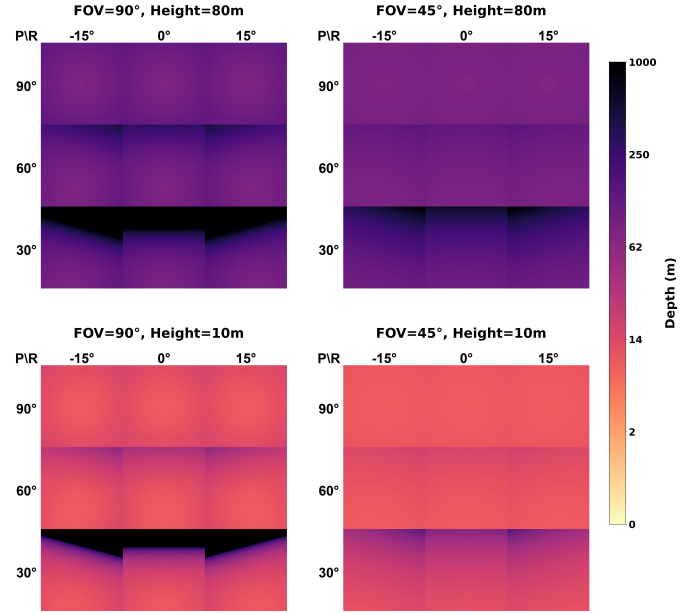}
\caption{Visualization of the Ideal Ground Depth distribution under varying camera parameters. This diagram illustrates the distinct effects of Pitch (P), Roll (R), Height, and FoV on the projected distance from the lens to the ground plane.}
\label{fig:d2233}
\end{figure}

Using the image extent $D$ and field of view ($\mathrm{fov}$) as shown in Fig.~\ref{fig:ll1}(b), the focal length $f$ is derived as:
\begin{equation}\label{eq:13}
\tan\left(\frac{\mathrm{fov}}{2}\right) = \frac{D}{2f} \implies f = \frac{D}{2\tan\left(\frac{\mathrm{fov}}{2}\right)}.
\end{equation}

\textbf{The FoV represents the angular extent of the observed image; by combining it with the image dimension $D$, the angular deviation of each pixel relative to the optical axis can be determined.} Substituting $f$ back into Eq.~\eqref{eq:12}, we obtain the specific calculation formulas:
\begin{equation}\label{eq:14}
\left\{
\begin{aligned}
\tan \theta_{x_I} &= \frac{2H_I \tan\left(\frac{\mathrm{fov}}{2}\right)}{H_{\max}}, \\
\tan \theta_{y_I} &= \frac{2W_I \tan\left(\frac{\mathrm{fov}}{2}\right)}{W_{\max}}.
\end{aligned}
\right.
\end{equation}

Finally, as illustrated in Fig.~\ref{fig:ll2}, let $h$ be the camera height and $\varphi$ be the pitch angle relative to the nadir. The ground-plane projection coordinates $(x_w, y_w)$ are given by:
\begin{equation}\label{eq:15}
x_{w} = h\,\tan\theta_{x_I},
\qquad
y_{w} = h\,\tan\left(\varphi + \theta_{y_I}\right).
\end{equation}

As seen from the expression for $y_w$, \textbf{the Pitch($\varphi$) determines the tilt of the optical axis, which leads to a non-linear expansion of depth in the upper regions of the image.} By utilizing $x_w$, $y_w$, and $h$, we can derive the distance $d_{up}$ from the lens to the target position:
\begin{equation}\label{eq:16}
d_{up} = \sqrt{x_{w}^2 + y_{w}^2 + h^2}.
\end{equation}

Expanding the above equation yields the following representation. It is evident that \textbf{the depth distance for all pixels is directly proportional to the Height ($h$), which acts as a linear scaling factor:}
\begin{equation}\label{eq:17}
d_{up} = h\,\sqrt{\tan^2\theta_{x_I} + \tan^2\left(\varphi + \theta_{y_I}\right) + 1}.
\end{equation}

\begin{figure*}[!t]
\centering
\includegraphics[width=0.85\linewidth]{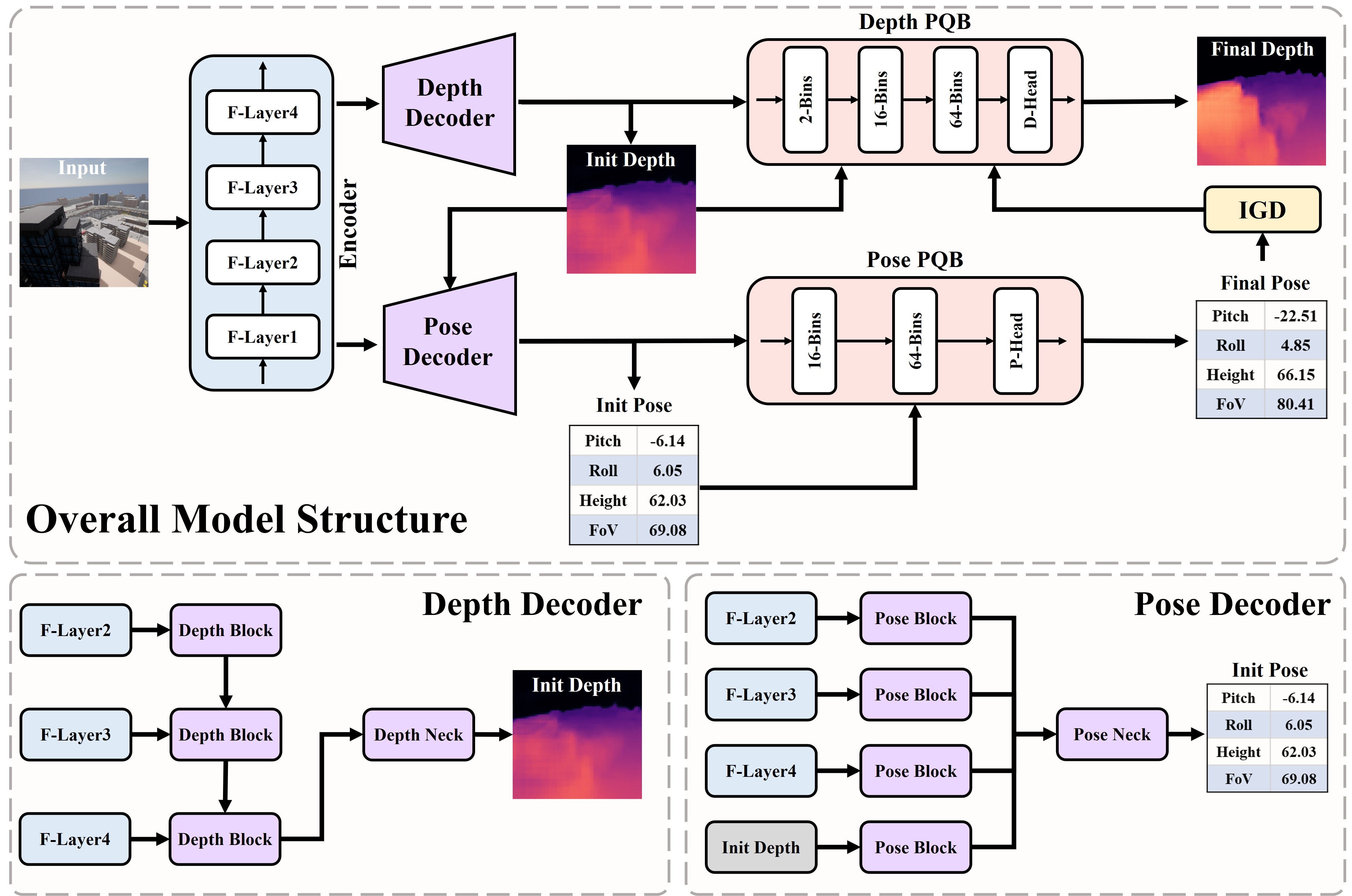}
\caption{Overall Model Structure. DAPM first extracts features using a shared encoder and then applies separate decoders for pose and depth to obtain initial estimates. It subsequently refines these predictions using the Progressive Quantization Bins and Ideal Ground Depth modules. DAPM demonstrates clear advantages in UAV imagery through its innovative design.}
\label{fig:ms}
\end{figure*}

Building upon this derivation, the parameter set $[\text{FoV}, \text{Pitch}, \text{Roll}, \text{Height}]$ governs the synthesis of the ideal ground depth. As visualized in Fig.~\ref{fig:d2233}, we analyze the isolated influence of each parameter on the depth map. While the ideal ground depth cannot serve as a rigid constraint due to real-world topographic undulations and pose estimation errors, it possesses significant theoretical value. First, it provides a geometric prior, offering a baseline context of sky and ground for aerial imagery. Second, the transformation relations indicate that the ideal ground depth explicitly embeds camera pose information, characterizing the imaging perspective. Consequently, we introduce the ideal ground depth as a feature input rather than a hard boundary, enhance the model's comprehension of the viewing angle, and improve depth estimation.

\subsection{Depth Estimation for Any Perspectives Model}

\subsubsection{Overall Model Architecture}


DAPM proposes a joint optimization strategy that prioritizes depth estimation while simultaneously incorporating camera pose estimation as an auxiliary task, tightly coupling the two through hierarchical feature sharing and task-specific constraints. As shown in Fig.~\ref{fig:ms}, the model first extracts multi-level features through a shared encoder. A decoding scheme is then applied to produce an initial depth prediction, which is progressively refined using a quantization-bin module to achieve high-accuracy results. The pose branch employs a simplified pipeline to minimize network capacity and computational cost, while maintaining the initial prediction and quantization stages. The Ideal Ground Depth module leverages the geometric consistency of the Ideal Ground Depth map to construct a pose loss, while simultaneously providing the sky-ground background prior features and viewing perspective representations to the depth quantization-bin module. DAPM employs a composite loss that jointly supervises quantized depth, quantized pose, continuous depth regression, and geometric consistency, thereby enhancing the overall performance of both depth and pose estimation.

\subsubsection{Encoder and Decoder}

\noindent \textbf{Encoder:} The encoder is based on a pre-trained ResNet50, from which multi-scale features are extracted from the second, third, and fourth residual blocks. These three sets of feature maps exhibit progressively decreasing spatial resolutions while capturing increasingly rich semantic information, thereby preserving fine details and incorporating high-level contextual cues. The extracted multi-scale features are subsequently fed into the DepthDecoder and PoseDecoder, respectively.
\\
\noindent \textbf{Depth Decoder:} Depth Decoder adopts a Feature Pyramid Network (FPN)-based architecture with residual enhancements to produce high-resolution depth maps from multi-scale encoder features. Specifically, lateral connections are constructed from three backbone stages via $1 \times 1$ convolutional layers and spatial dropout to enhance feature selectivity and regularization. These enhanced lateral features are then fused in a top-down manner using residual convolutional blocks to preserve semantic richness while progressively increasing spatial resolution. At each stage of fusion, lower-resolution features are upsampled and combined with higher-resolution lateral features, allowing for multi-scale integration with residual refinement. Finally, the aggregated high-resolution feature is further processed through a residual prediction head, followed by a $1 \times 1$ convolution and a sigmoid activation to generate the normalized depth map. The output includes both the final upsampled feature representation and the predicted dense depth map, supporting accurate and robust depth estimation from monocular input.
\\
\noindent \textbf{Pose Decoder:} The Pose Decoder is designed to estimate camera pose parameters while jointly learning multi-scale quantization representations. Since pose estimation predicts a 1×4 vector that reflects the global characteristics of the entire image, shallow features are more effective than deeper ones for capturing such global structural cues. Therefore, instead of only relying primarily on deep features, we directly concatenate multi-level features from different encoder stages to better exploit the macroscopic information preserved in shallow layers. The decoder first aggregates these multi-scale features using global average pooling and linear projections to form a unified pose representation. The fused representation is then processed by a residual MLP to produce an initial 4D pose vector, followed by a sigmoid activation.


\subsubsection{Ideal Ground Depth}

\begin{figure}[t]
\centering
\includegraphics[width=0.9\linewidth]{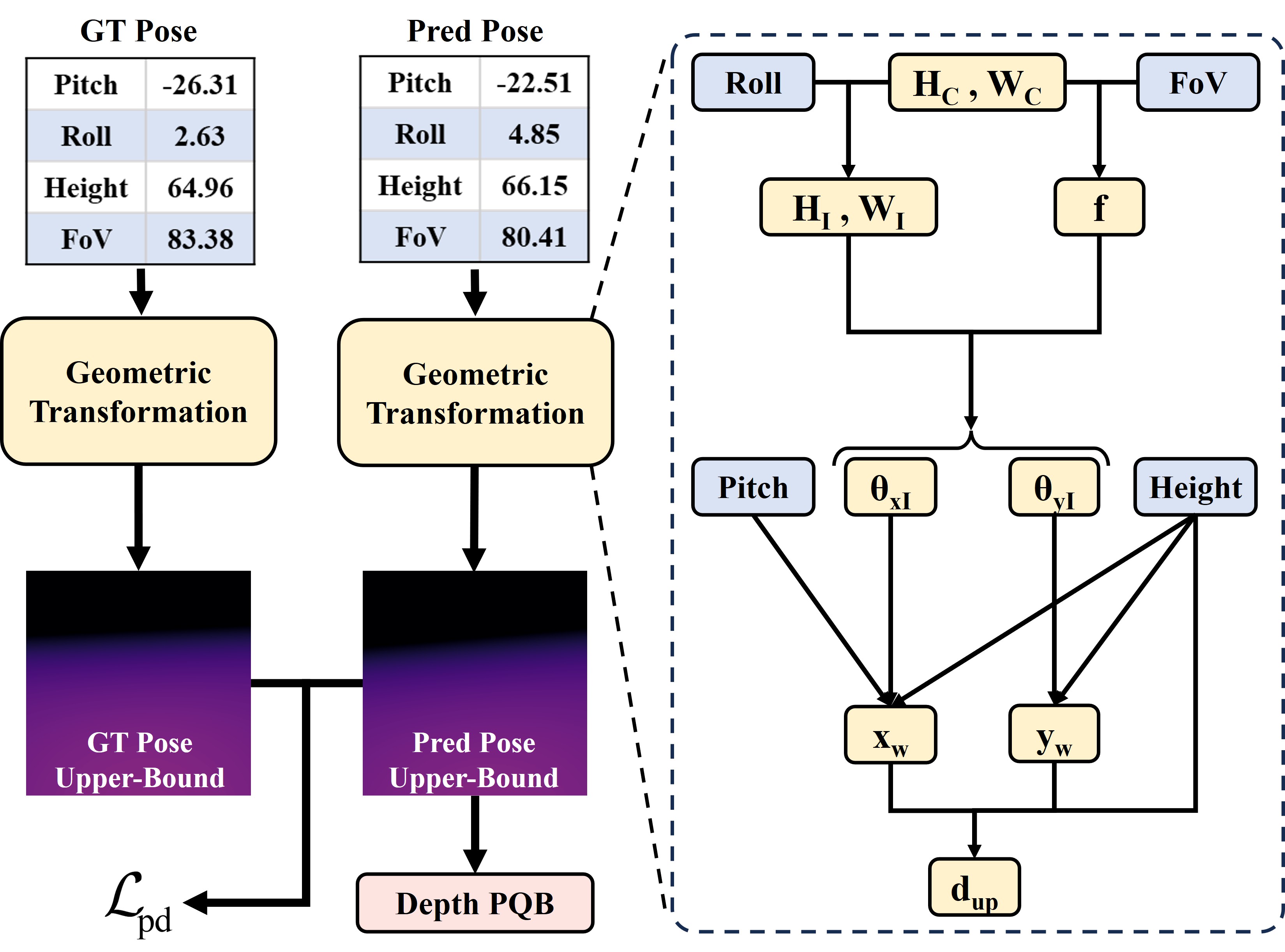}
\caption{The IGD follows the methodology described in \ref{subsec:addgo} to perform an Ideal Ground Depth conversion on both the estimated and ground-truth pose values, and computes the loss between the converted results. Specifically, the geometric transformation process is carried out by first obtaining the angular deviation between each pixel and the optical axis, then projecting into the transformed world coordinate system, thereby obtaining the ground coordinates corresponding to the pixel, and finally calculating the distance between the ground-plane position corresponding to each pixel and the camera in the world coordinate system.}
\label{fig:dub}
\end{figure}

To effectively leverage the ideal ground depth distribution as described in Section \ref{subsec:addgo} for enhancing the performance of our model, we propose two complementary strategies that target both pose and depth estimation tasks. As shown in Fig.~\ref{fig:dub}, the first strategy involves the introduction of an Ideal Ground Depth-guided loss to improve pose estimation. Specifically, we transform both the ground-truth and predicted camera poses into their corresponding ideal ground depth maps. By computing the mean squared error (MSE) between these maps, pose regression is reformulated from conventional vector-based supervision into dense matrix-level comparison. This transformation allows the network to receive richer and spatially detailed gradient information during training, thereby facilitating more effective learning of pose features and capturing subtle geometric cues that may be overlooked under standard vector supervision.

The second strategy aims to enhance depth estimation by integrating the ideal ground depth distribution into the multi-level Bins Blocks of the depth estimation network. The ideal ground depth maps, derived from the estimated pose information, are subsequently processed through convolutional layers and concatenated with intermediate feature representations. This design enables the model to explicitly incorporate background prior features composed of the ideal sky and ground, along with perspective representations of the viewing angle, thereby enhancing the model's capability. By fusing the ideal ground depth distribution with multi-level feature representations, the network is provided with an additional structured prior that complements the learned visual features, reinforcing the consistency between estimated poses and corresponding depth values. Functioning simultaneously as a dense supervisor for pose estimation and a geometric prior for depth estimation, the Ideal Ground Depth module leverages the intrinsic pose-depth correlation to mutually reinforce both tasks. This synergy not only improves estimation accuracy and robustness against complex geometries but also provides a principled mechanism for the rigorous joint optimization of pose and depth.

\subsubsection{Progressive Quantization Bins}

\begin{figure*}[!htb]
\centering
\includegraphics[width=0.85\linewidth]{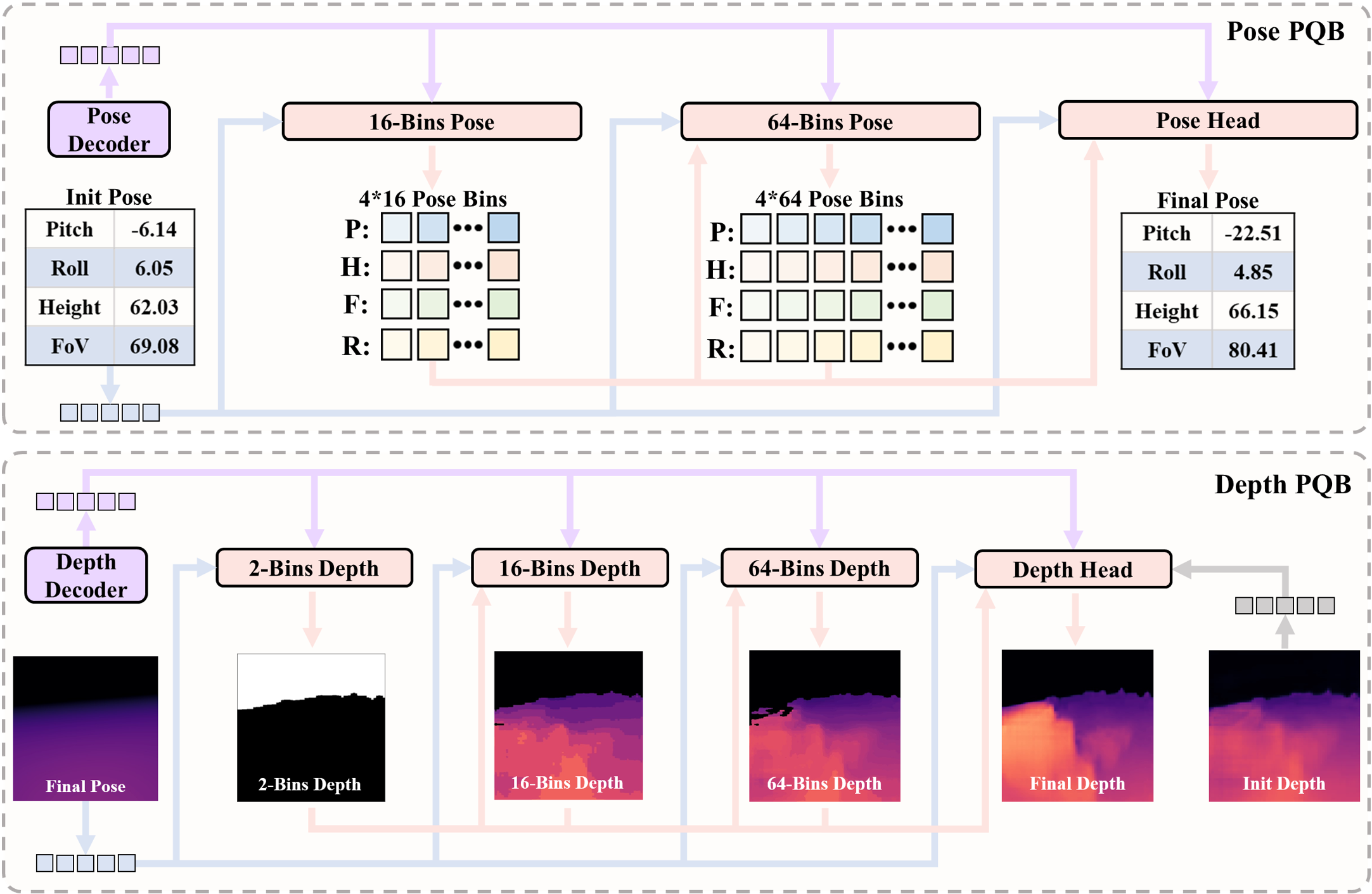}
\caption{Progressive Quantification Bins Module. This module gradually improves the estimation accuracy of camera pose and depth by continuously increasing the number of bins for classification estimation. Each bin block obtains the corresponding classification result by concatenating the input features and performing convolution processing. Finally, the depth head and pose head fuse all the preceding features to give the final estimation result.}
\label{fig:quant}
\end{figure*}

To achieve a coarse-to-fine progressive estimation, as illustrated in Fig.~\ref{fig:quant}, the Progressive Quantization Bins Module consists of two main components: Pose Progressive Quantification Bins (Pose PQB) and Depth Progressive Quantification Bins (Depth PQB). The Pose PQB component takes the features extracted by the Pose Decoder, together with the initial depth estimates as input, and performs a progressive quantization of the camera pose. Specifically, the pose parameters are first classified into 16-bin discrete intervals and are then refined into 64-bin intervals, progressively narrowing the estimation range until the final pose prediction is obtained. The quantized results from the preceding stage are also retained and processed as auxiliary features to guide the subsequent stage of estimation. This iterative refinement strategy enhances the precision of pose estimation and maintains stability in the optimization process.

The Depth PQB component adopts a similar progressive quantization structure. Unlike previous methods that construct unified bins for all pixels in the entire image, we construct bins for each pixel individually. This effectively avoids the conflicting effects of pixels with different depth scales on the global bins, thereby enhancing depth estimation performance under the large-scale depth distributions of UAV imagery. However, considering that aerial imagery often contains a large proportion of pixels corresponding to regions such as the sky that extend beyond the effective depth range, an additional enhancement is introduced. In addition to the standard bins classification, a binary classification function is incorporated to determine whether a pixel exceeds the defined depth threshold. Furthermore, the Ideal Ground Depth Map refined by the Ideal Ground Depth Module is integrated as a feature in each estimation layer, providing additional geometric feature enhancement and facilitating more reliable depth estimation across different distance scales. Within each Bins Block, the process begins with feature concatenation, followed by two consecutive convolutional layers that effectively integrate multi-scale spatial information.

The final results are obtained through two specialized task heads; the Depth Head is composed of multiple convolution and Batch Normalization layers to produce refined depth quantization outputs. The Pose Head has a similar architecture but replaces the convolutional layers with Linear Refinement Layers, which are more suitable for processing the global pose parameters. Through this hierarchical and progressive quantization mechanism, the Progressive Quantization Bins Module enables the model to achieve a coarse-to-fine estimation process. This design improves both the accuracy and robustness of depth and pose estimation, particularly under complex aerial imaging conditions.

\subsubsection{Loss Function}

In this work, we propose a composite loss function within a multi-task learning framework for simultaneous depth estimation and pose regression. First, the binary depth branch and the quantized depth branches are supervised by a combination of binary cross-entropy and multi-class cross-entropy losses, yielding $\mathcal{L}_{\mathrm{dq}}$:

\begin{equation}\label{eq:16}\tag{16}
\begin{aligned}
\mathcal{L}_{\mathrm{dq}} = & \mathcal{L}_{\mathrm{BCE}}\left(\hat{D}_{\mathrm{q2}}, D_{\mathrm{q2}}\right) 
+ \mathcal{L}_{\mathrm{CE}}\left(\hat{D}_{q16}, D_{q16}\right) \\
& + \mathcal{L}_{\mathrm{CE}}\left(\hat{D}_{q64}, D_{q64}\right).
\end{aligned}
\end{equation}


Second, for both the initial and the refined depth predictions, we employ a scaled variant of the Scale-Invariant loss introduced by Eigen\cite{eigen2014depth} et al.:

\begin{equation}\label{eq:17}\tag{17}
\mathcal{L}_{\text{reg}} = \alpha \sqrt{
  \frac{1}{T}\sum_{i}g_{i}^{2}
  - \frac{\lambda}{T^{2}}\left(\sum_{i}g_{i}\right)^{2}
},
\end{equation}


Where $g_{i} = \log\tilde{d}_{i} - \log d_{i}$ with the ground truth depth $d_{i}$ and predicted depth $\tilde{d}_{i}$, $T$ denotes the number of pixels having valid ground truth values, and we set $\lambda = 0.85$ and $\alpha = 10$ following Adabins. We refer to the losses on the initial and refined depth outputs as $\mathcal{L}_{\text{init-depth}}$ and $\mathcal{L}_{\text{depth}}$, respectively.

For the pose estimation branch, the quantized pose components are supervised by multi-class cross-entropy, yielding $\mathcal{L}_{\mathrm{pq}}$:

\begin{equation}\label{eq:18}\tag{18}
\mathcal{L}_{\mathrm{pq}} = \mathcal{L}_{\mathrm{CE}}\left(\hat{P}_{q16}, P_{q16}\right) 
+ \mathcal{L}_{\mathrm{CE}}\left(\hat{P}_{q64}, P_{q64}\right).
\end{equation}


While the initial and refined pose predictions are optimized with mean squared error losses $\mathcal{L}_{\text{init-pose}}$ and $\mathcal{L}_{\text{pose}}$. Furthermore, to ensure geometric consistency, we introduce the dense pose supervision loss $\mathcal{L}_{\mathrm{pd}}$ derived from the IGD. To balance the weights of the different losses, we assign $\rho=10$ to ensure that all loss values are of the same order of magnitude. Finally, the overall training objective is defined as the weighted sum of all sub-losses, integrating multiple loss components to facilitate more effective model learning during training:

\begin{equation}\label{eq:19}\tag{19}
\begin{aligned}
\mathcal{L}_{\text{total}} = & \left(\mathcal{L}_{\text{depth}} + \rho\mathcal{L}_{\text{pose}}\right) 
+ \left(\mathcal{L}_{\text{init-depth}} + \mathcal{L}_{\text{init-pose}}\right) \\
& + \mathcal{L}_{\mathrm{dq}} + \mathcal{L}_{\mathrm{pq}} + \rho\mathcal{L}_{\mathrm{pd}}.
\end{aligned}
\end{equation}


\begin{table*}[t]
\centering
\caption{Comparison of UAPD with Other Drone Depth Estimation Datasets}
\label{tab:uav_datasets}
\begin{threeparttable}
\setlength{\tabcolsep}{4pt}
\renewcommand{\arraystretch}{1.1}
\resizebox{0.8\linewidth}{!}{%
\begin{tabular}{l c c c c c}
\toprule
Name & Depth [m] & Height [m] & Pitch [deg] & Fov [deg] & Roll [deg] \\
\hline
Mid-Air\tnote{*} \cite{fonder2019mid} & -- & [-500,500] & [-30,25] & 90 & [-35,35] \\
SynDrone \cite{rizzoli2023syndrone} & [0,1000] & 20, 50, 80 & 30, 60, 90 & 90 & -- \\
SkyScenes\tnote{*} \cite{khose2024skyscenes} & -- & 15, 35, 60 (±2.5) & 0, 45, 60, 90 (±2.5) & 110 & -- \\
DDOS \cite{kolbeinsson2024ddos} & [0,100] & [1,25] & [-45,30] & -- & [-40,40] \\
UEMM-Air\tnote{*} \cite{yao2025uemmair} & [0,100] & [5,50] / 5 & [0,90] / 5 & -- & -- \\
\hline
UAPD (Ours) & [0,1000] & [10,100] & [-10,100] & [30,120] & [-30,30] \\
\toprule
\end{tabular}
}%
\begin{tablenotes}\footnotesize
\begin{minipage}{1.55\linewidth}
\item  \tnote{*}  Mid-Air: 'Height \(<\) 0' is due to the author not setting the ground plane as the lowest. SkyScenes: '±2.5' indicates that the interval is 2.5 above and below each value. UEMM-Air: '/ 5' indicates that the sampling interval within the interval is 5.
\end{minipage}
\end{tablenotes}
\end{threeparttable}
\end{table*}

\begin{figure*}
\centering
\includegraphics[width=0.8\linewidth]{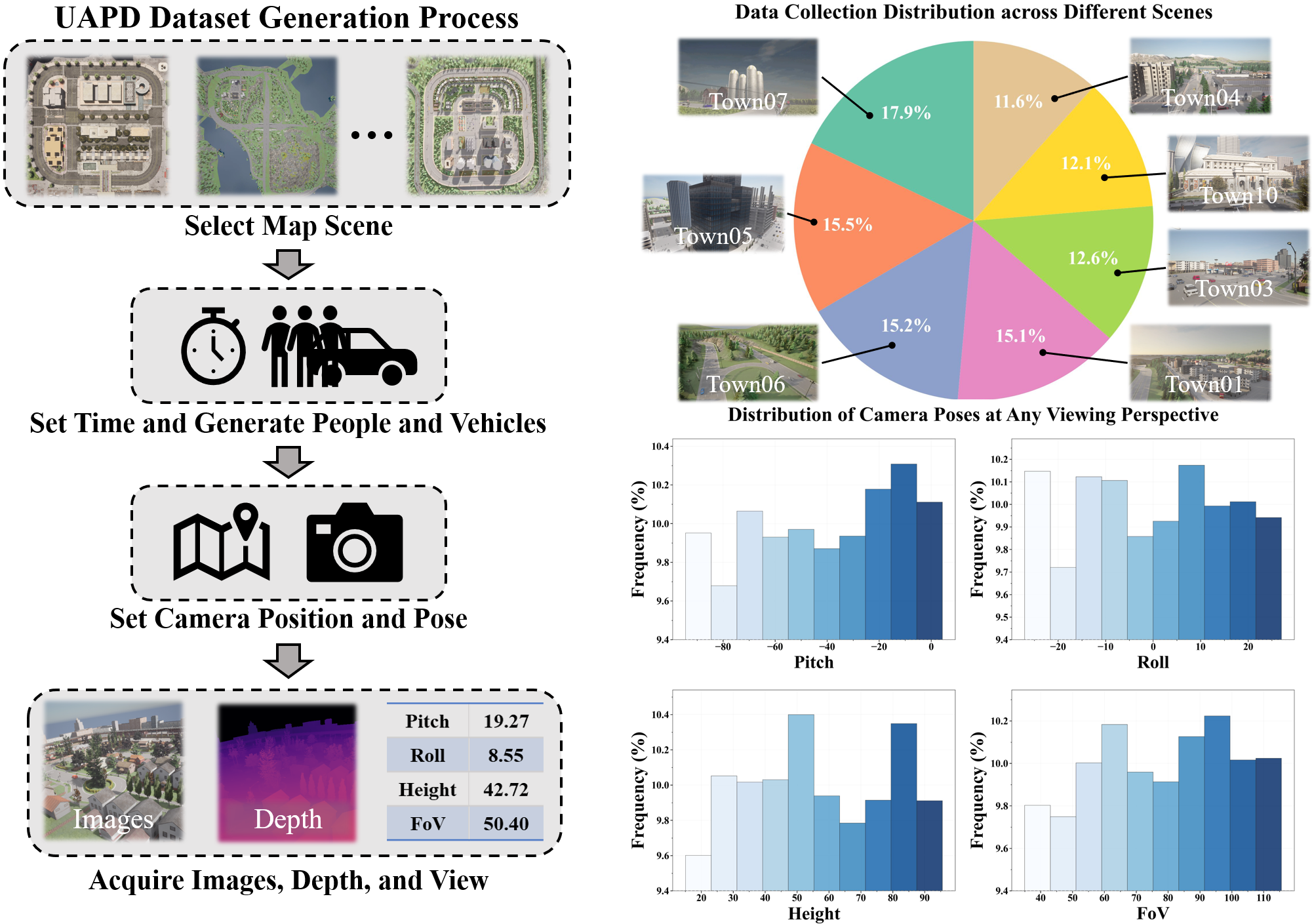}
\caption{By selecting simulation scenes, setting the time, and generating random pedestrians and vehicles, a realistic simulation environment is constructed. Subsequently, images and their corresponding depth maps are acquired by randomly sampling camera positions and poses. In addition, we provide the data proportions of different scenes and the distribution of viewing perspectives within the UAPD dataset.}
\label{fig:uapd}
\end{figure*}

\subsection{UAV Any Perspectives Depth dataset}
\subsubsection{Dataset Overview}

To evaluate the model's depth and camera pose estimation capabilities on UAV imagery with diverse viewpoint distributions, we propose the UAPD dataset. UAPD was collected using CARLA simulation data and contains a total of 42k images, along with corresponding depth maps and parameter information. Both the images and depth maps have a resolution of 512x512. During data collection, we randomly set lighting conditions, camera poses, and the distribution of vehicles and pedestrians across various environments, including urban, rural, and agricultural settings, effectively enhancing the diversity of the data. The UAPD consists of two parts: a training set and a validation set. The training set contains 41,186 images, and the validation set contains 857 images. As shown in Tab.~\ref{tab:uav_datasets}, compared with other datasets, UAPD has a wide range of continuous distributions in depth, pitch, roll, height, and FoV. Therefore, UAPD effectively characterizes the image performance and depth distribution of drone images at any viewing angle.

\subsubsection{Data Acquisition}

As shown in Fig.~\ref{fig:uapd}, we developed a configurable CARLA-based data acquisition system for large-scale training of autonomous driving perception models. The system dynamically loads one of seven town maps. Note that data collection was not performed in Town02 due to its high similarity to Town01. Weather conditions are fixed to ClearNoon, while the time of day is randomly selected between 8:00 and 17:00 to simulate varying solar elevation angles. After each reset, 200 vehicles and 200 pedestrians with randomized behaviors are spawned to construct complex dynamic environments. Camera poses are randomized within a $100 \times 100$ m area centered at the origin. To simulate real-world optical distortions, physical lens properties are randomized, with specific parameter settings and distribution ranges as summarized in Tab.~\ref{tab:param_ranges_v2}. All effects are rendered in real time via the CARLA physics engine to ensure photorealism. The main acquisition loop executes 50k independent sampling iterations under system control. After each acquisition cycle, a 200-millisecond sleep interval is implemented to guarantee stable data writing. Then we manually removed 8k images that had modding issues due to random camera positions.



\begin{figure}
\centering
\includegraphics[width=0.9\linewidth]{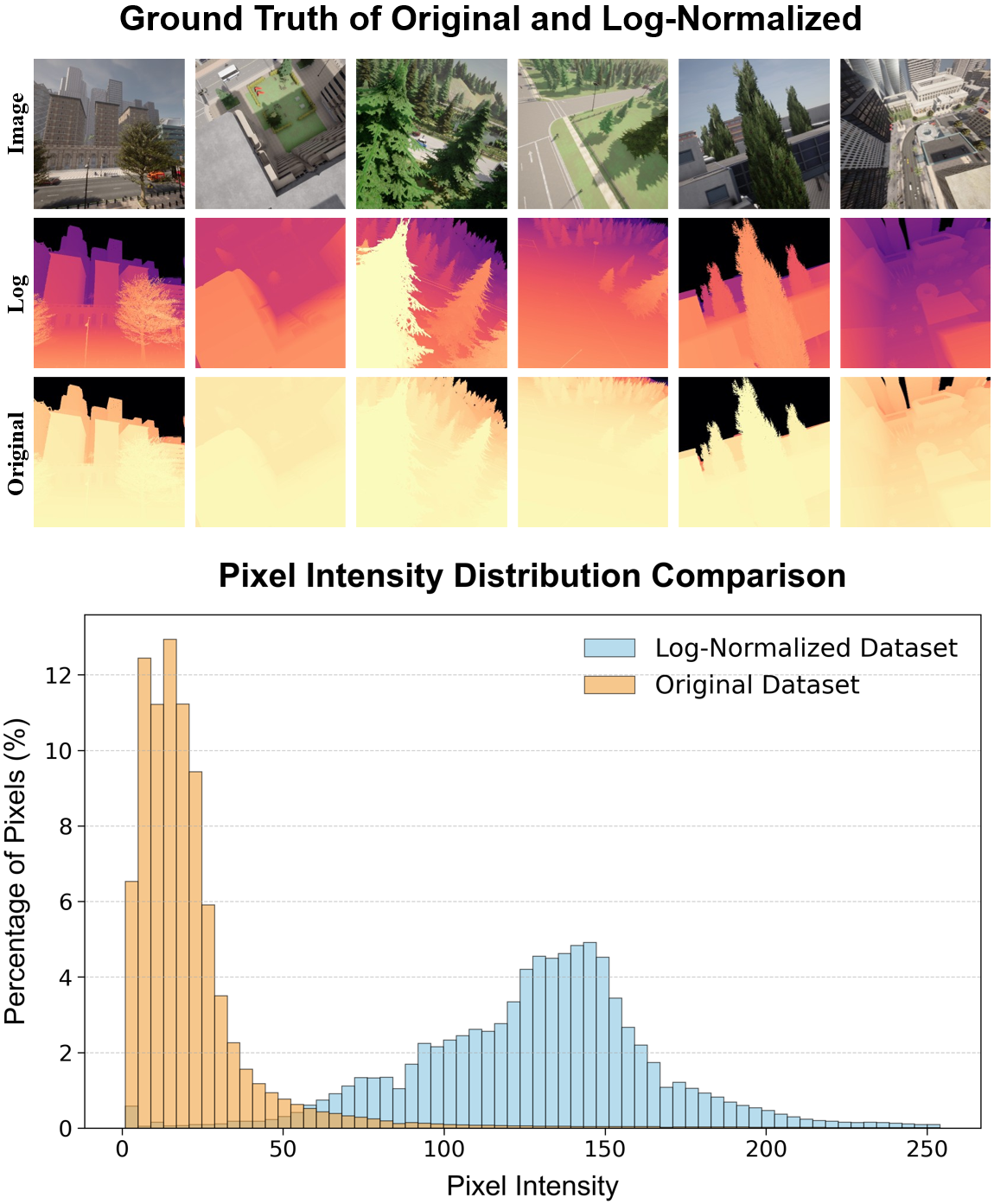}
\caption{Through qualitative visualization of the ground-truth depth maps and comparative analysis of the depth histogram distributions between the original and log-normalized depth maps, it can be observed that the log-normalized depth maps exhibit superior visual quality. Moreover, their depth distributions align more closely with a Gaussian distribution, providing a more effective representation of the scene depth characteristics.}
\label{fig:his}
\end{figure}

When obtaining depth maps, we can choose to obtain the original depth map directly from linear mapping, or we can choose the log-normalized depth map. As shown in Fig. \ref{fig:his}, the depth map distribution of the dataset obtained directly from linear mapping has a significant long-tail effect, while the log-normalized distribution is closer to a Gaussian distribution. Therefore, we choose the log-normalized depth map as the depth ground truth for the dataset. The depth map's range corresponds to 0–1,000 meters, with a logarithmic relationship to grayscale values, as detailed below:


\begin{equation}\label{eq:20}\tag{20}
\text{d}_{\log} = 255 \cdot \frac{\log_2(\text{d} + 1)}{\log_2(\text{d}_{\max} + 1)}.
\end{equation}

Where d is the actual distance between the lens and the corresponding pixel location, and d\_{max} is the maximum distance of 1000m. When the distance exceeds 1000m, it is represented as a truncated value of 1000m. The above formula first applies logarithmic scaling to the distance, then normalizes it by comparing it to the maximum value, and finally linearly scales it to the 0-255 grayscale value range. This dataset provides substantial data support for subsequent research tasks involving UAV camera pose and image depth estimation.




\begin{table}[t]
\centering
\caption{Camera Parameter Ranges of the UAPD Dataset}
\label{tab:param_ranges_v2}
\begin{threeparttable}
\resizebox{1\linewidth}{!}{%
\begin{tabular}{ccc ccc}
\toprule
\textbf{Param} & \textbf{Min} & \textbf{Max} & \textbf{Param} & \textbf{Min} & \textbf{Max} \\
\midrule
X(m) & -50 & 50 & Z(m)\tnote{*} & 10 & 100 \\
Y(m) & -50 & 50 & FoV(deg) & 30 & 120 \\
Pitch(deg) & -100 & 10 & Yaw(deg) & -180 & 180 \\
Roll(deg) & -30 & 30 & Lens kcube & -0.2 & 0.2 \\
Time(h) & 8 & 17 & Lens k & -0.5 & 0.5 \\
Lens X Size & 0.3 & 0.7 & Lens Y Size & 0.3 & 0.7 \\
\toprule
\end{tabular}
}
\begin{tablenotes}\footnotesize
\begin{minipage}{1.55\linewidth}
\item \tnote{*} The Z represents height.
\end{minipage}
\end{tablenotes}
\end{threeparttable}
\end{table}

\section{Experiments}

\subsection{Implementation Details and Evaluation Metrics}

\noindent \textbf{Training Setup.} Our implementation is based on the PyTorch framework. The entire model is trained for 32k iterations on a single node equipped with four NVIDIA A40 GPUs, using a batch size of 32 and an initial learning rate of $1\times 10^{-4}$. We adopt the AdamW optimizer with parameters $(\beta_1, \beta_2, \textit{wd}) = (0.9, 0.999, 0.01)$, where \textit{wd} denotes the weight decay. A linear warm-up strategy is applied during the first 30\% of the training iterations. For learning rate scheduling, we employ the cosine annealing strategy. Due to the limited computational power of UAV platforms, backbones with massive parameters are difficult to deploy in practice; therefore, to achieve a balance between model accuracy and inference speed, our method utilizes the standard convolutional ResNet-50 \cite{he2016deep} backbone. Due to the low accuracy of model-estimated camera poses during the early stages of training, the pose input to the IGD module is derived from a weighted sum of the ground truth and the model estimation. We employ a dynamic weight adjustment strategy that linearly increases the proportion of the estimated values as the number of epochs increases. Consequently, the model transitions to using the estimated values entirely by the end of training and during inference, thereby enhancing training stability and convergence robustness.
\\
\noindent \textbf{Evaluation Metrics.} We follow the standard evaluation protocol adopted in previous studies \cite{veicht2024geocalib} \cite{li2024binsformer} to validate the effectiveness of our method in the experiment. Specifically, for depth estimation, we use threshold-based accuracy ($\delta_i < 1.25^i$, $i=1,2,3$), average absolute relative error (AbsRel), root mean square error (RMSE), and root mean square $\log_{10}$ error (log10). For each metric in the camera pose estimation task, we report the median error and area under the recall curve (AUC) up to 1/5/10°.
\\
\noindent \textbf{Data Processing and Augmentation.} During dataset preprocessing, the ground truth depth values are first log-transformed, and the labels are normalized to ensure that all parameters fall within the $[0, 1]$ range. To enhance the model's ability to learn from discretized representations, both depth maps and pose parameters undergo uniform quantization followed by one-hot encoding. For data augmentation, a lightweight image transformation pipeline is employed, including color perturbations (brightness, contrast, and saturation variations) and the addition of Gaussian noise. Random horizontal flipping is also applied, with synchronized adjustment of the roll component in the pose vector to maintain geometric consistency. Vertical flipping and random rotations are avoided, as they may introduce inconsistencies in the pose characteristics of the images.

\begin{table*}[t]
\centering
\caption{Comparison of Depth Estimation Performance on UAPD Dataset}
\begin{threeparttable}
\setlength{\tabcolsep}{4pt}
\renewcommand{\arraystretch}{1.1}
\resizebox{0.6 \linewidth}{!}{%
\begin{tabular}{lcccccc}
\toprule
\textbf{Method} & $\delta_1 \uparrow$ & $\delta_2 \uparrow$ & $\delta_3 \uparrow$ & AbsRel $\downarrow$ & RMS $\downarrow$ & $log_{10} \downarrow$ \\
\midrule
BTS \cite{lee2019big} & 0.864 & 0.962 & 0.979 & 0.376 & 0.154 & 0.067 \\
DepthFormer \cite{li2023depthformer} & 0.893 & 0.977 & 0.983 & 0.314 & 0.145 & 0.047 \\
NeWCRFs \cite{yuan2022neural} & 0.895 & 0.976 & 0.984 & 0.296 & 0.131 & 0.052 \\
DPT \cite{ranftl2021vision} & 0.883 & 0.969 & 0.985 & 0.358 & 0.133 & 0.056 \\
Adabins \cite{bhat2021adabins} & 0.836 & 0.956 & 0.981 & 0.322 & 0.160 & 0.063 \\
Simipu \cite{li2022simipu} & 0.877 & 0.970 & 0.985 & 0.271 & 0.139 & 0.055 \\
BinsFormer \cite{li2024binsformer} & 0.899 & 0.973 & 0.987 & 0.281 & 0.129 & 0.053 \\
\midrule
DAPM(D)\tnote{*} & 0.916 & 0.975 & 0.986 & 0.274 & 0.148 & 0.041 \\
\textbf{DAPM} & \textbf{0.927} & \textbf{0.983} & \textbf{0.989} & \textbf{0.252} & \textbf{0.103} & \textbf{0.034} \\
\bottomrule
\end{tabular}
}
\begin{tablenotes}\footnotesize
\begin{minipage}{2\linewidth}
\item  \tnote{*}DAPM(D) indicates that the model is trained only for depth estimation, while DAPM is trained jointly for camera pose and depth estimation.
\end{minipage}
\end{tablenotes}

\end{threeparttable}
\label{tab:table4}
\end{table*}

\subsection{Comparison with Monocular Depth Estimation Models on UAPD Dataset}
\label{sec:DEE}




To comprehensively evaluate the effectiveness of recent Monocular Depth Estimation (MDE) methods on the proposed UAPD dataset, we employed the Monocular Depth Estimation Toolbox \cite{lidepthtoolbox2022} for standardized training and evaluation, ensuring a fair comparison under identical experimental settings. All models utilized ResNet-50 as the backbone to maintain architectural consistency. The quantitative results are summarized in Tab.~\ref{tab:table4}.

\begin{figure*}[t]
\centering
\includegraphics[width=0.90\linewidth]{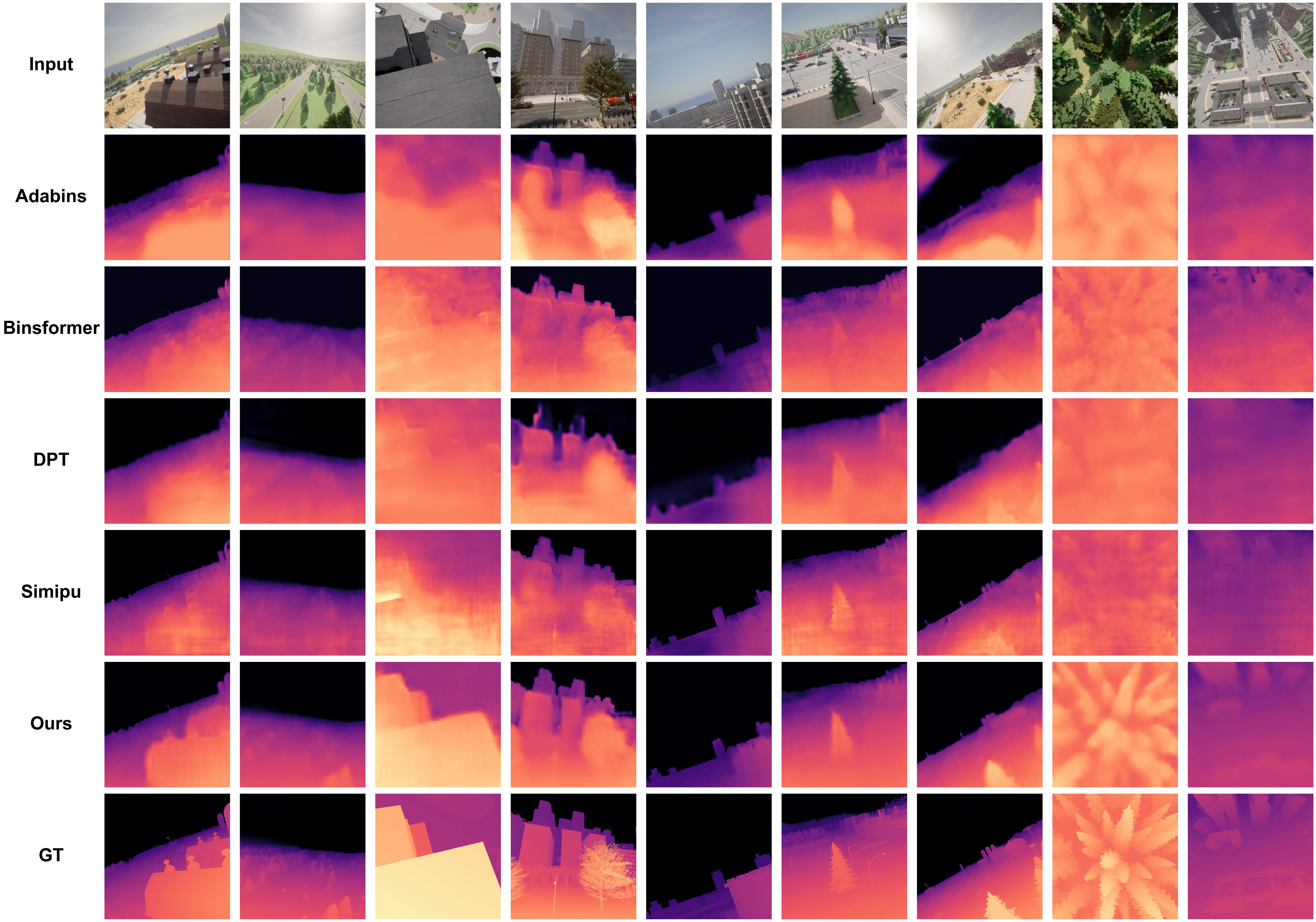}
\caption{Qualitative comparison of different depth estimation methods and the DAPM on the UAPD dataset. The results reveal that DAPM consistently produces more accurate depth estimates across a wide range of viewing angles, which clearly demonstrates its significant advantages over existing methods.}
\label{fig:vis}
\end{figure*}

As shown in Tab.~\ref{tab:table4}, the proposed DAPM model achieves state-of-the-art (SOTA) performance across all evaluation metrics, demonstrating clear advantages over existing methods. Specifically, the DAPM(D) variant, trained solely for depth estimation, already outperforms existing state-of-the-art methods such as BinsFormer on key metrics like $\delta_1$, fully confirming that the foundational depth network architecture proposed herein possesses superior feature extraction and regression capabilities. Meanwhile, compared to DAPM(D), the complete DAPM model employing the joint training strategy achieves a significant leap in performance, effectively demonstrating that incorporating the camera pose estimation task provides critical geometric features that further calibrate and optimize depth prediction results. In addition, qualitative comparisons illustrated in Fig.~\ref{fig:vis} further validate the quantitative findings. The DAPM model produces sharper object boundaries, more consistent depth transitions, and fewer artifacts in areas. Overall, both the quantitative and qualitative results demonstrate that DAPM not only achieves superior numerical accuracy but also provides visually coherent and structurally consistent depth predictions on aerial scenes, highlighting its potential for UAV-based perception and 3D reconstruction applications.

\begin{table}[t]
\centering
\caption{Comparison of Model Efficiency among Different Depth Estimation Methods}
\resizebox{0.75\linewidth}{!}{%
\begin{tabular}{lcc}
\toprule
\textbf{Method} & \textbf{Params (M)} & \textbf{FPS} \\
\midrule
BinsFormer \cite{li2024binsformer} & 81.96 & 11.5 \\
AdaBins \cite{bhat2021adabins} & 83.91 & 14.8 \\
Simipu \cite{li2022simipu} & 77.82 & 15.3 \\
DPT \cite{ranftl2021vision} & 109.99 & 8.6 \\
\midrule
\textbf{DAPM} & \textbf{42.82} & \textbf{24.1} \\
\bottomrule
\end{tabular}
}
\label{tab:efficiency}
\end{table}

A quantitative comparison of model efficiency is essential to evaluate the trade-off between computational complexity and real-time performance in depth estimation. Such analysis is particularly important for applications deployed on resource-constrained platforms, including unmanned aerial vehicles (UAVs) and mobile devices. As shown in Tab.~\ref{tab:efficiency}, DAPM demonstrates clear advantages in both model compactness and speed. With only 42.82 million parameters, it is much lighter than BinsFormer, AdaBins, Simipu, and DPT, while achieving the fastest inference speed of 24.1 FPS. These results highlight DAPM’s strong balance between efficiency and performance, enabling real-time operation with reduced memory and computation demands. This makes DAPM a practical and scalable solution for real-world depth estimation tasks in aerial and embedded vision systems.

\begin{figure*}
\centering
\includegraphics[width=1\linewidth]{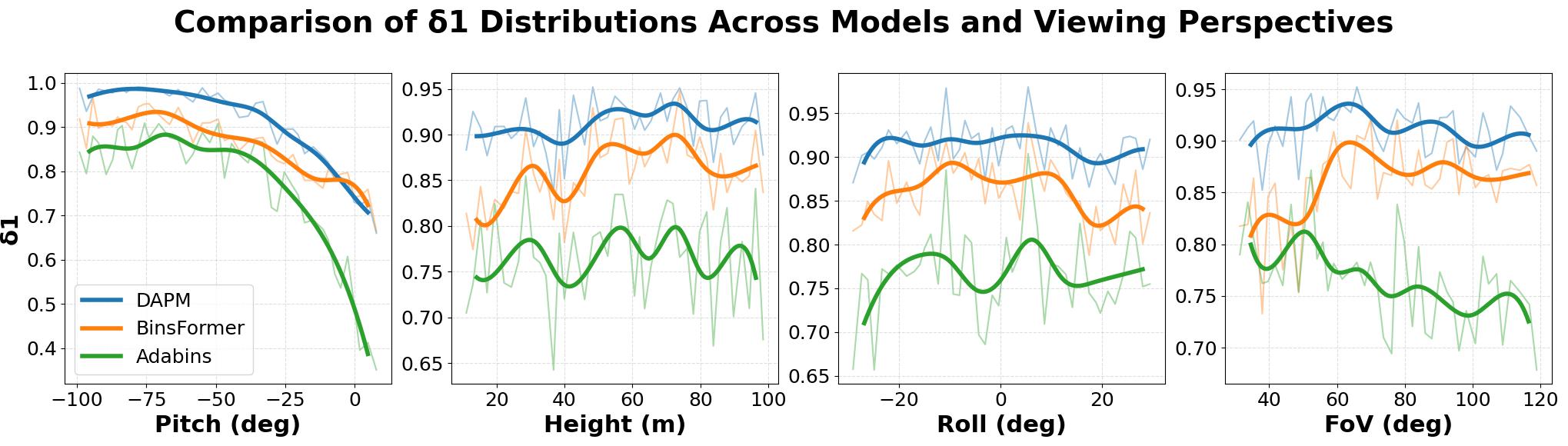}
\caption{We presents the accuracy performance of various depth estimation methods across different viewpoint distributions within the UAPD dataset. It can be seen that under all perspectives, DAPM demonstrates more stable and accurate performance compared to other methods.}
\label{fig:view_vis}
\end{figure*}

To analyze the impact of viewing angle variations on model performance, Fig.~\ref{fig:view_vis} compares the performance distributions of DAPM, BinsFormer, and AdaBins under four different viewing parameters: Pitch, Height, Roll, and Field of View (FoV). Overall, DAPM consistently outperforms the other two models across all viewing conditions, with its $\delta_1$ curve remaining the highest throughout, demonstrating superior robustness and overall accuracy. Among these parameters, Pitch variation has the most pronounced influence on $\delta_1$ for all models. As the pitch angle changes from -100° to 0°, the $\delta_1$ values show a clear downward trend, indicating a gradual performance decline at steeper viewing angles. In contrast, the distributions for Height, Roll, and FoV exhibit more fluctuations rather than a monotonic trend, suggesting that the relationship between these parameters and model performance is more complex and nonlinear. Nevertheless, DAPM exhibits notably more stable performance under varying viewing conditions compared with other depth estimation methods, confirming its strong robustness and adaptability in depth estimation across diverse camera perspectives.

\begin{figure}[t]
\centering
\includegraphics[width=0.95\linewidth]{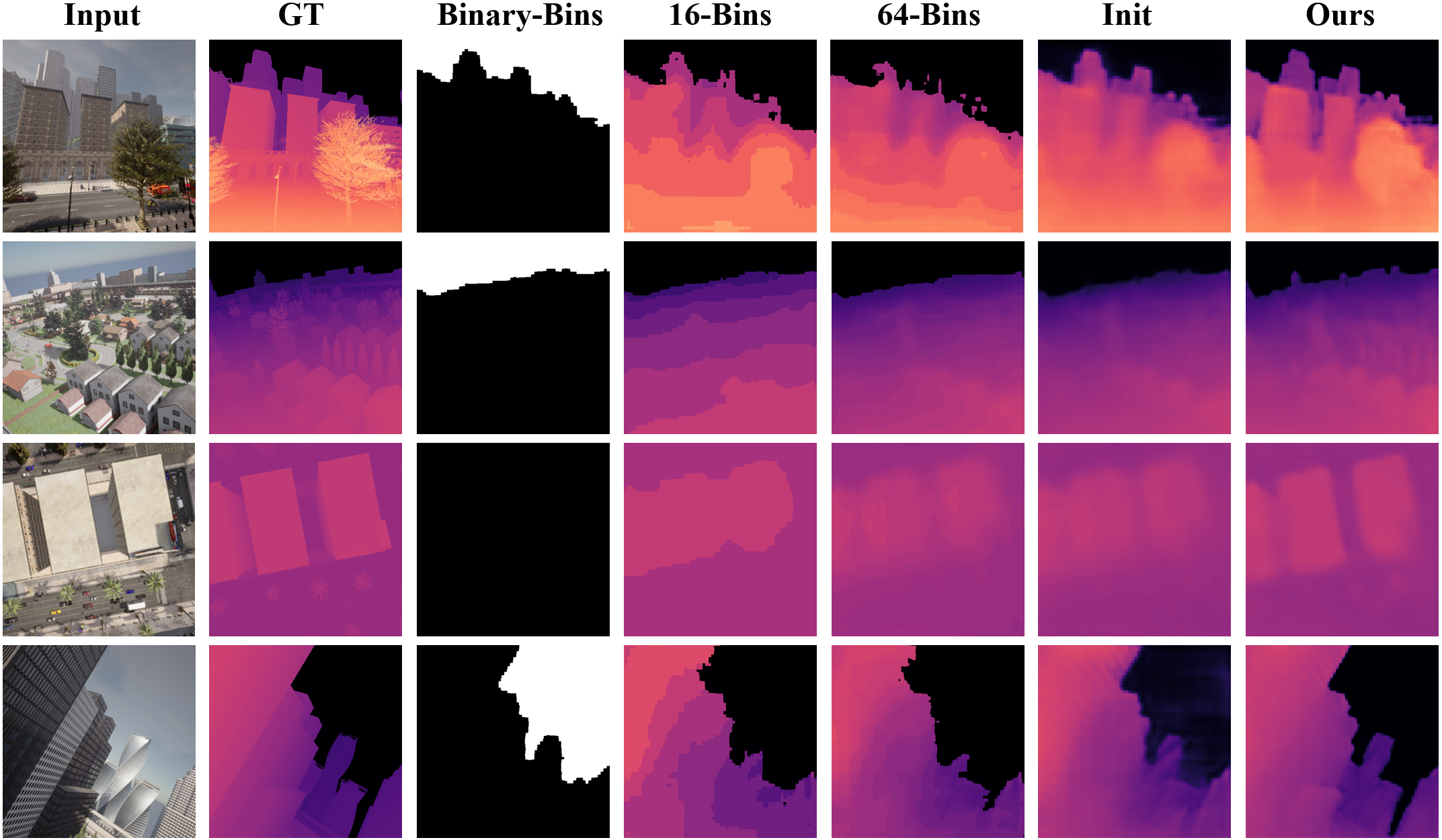}
\caption{The visualization of different quantization bins in the Depth PQB on the UAPD dataset clearly demonstrates that as the number of bins increases, the depth maps become progressively more refined. Furthermore, after applying the PQB enhancement, the final model output exhibits significantly higher accuracy compared to the initial prediction results.}
\label{fig:pqb_vis}
\end{figure}

Furthermore, to intuitively illustrate the improvements achieved by the PQB method, Fig.~\ref{fig:pqb_vis} provides a visual exposition of the depth estimation process. First, the model segments regions beyond the depth threshold (e.g., sky) through binary classification. Subsequently, the 16-Bins and 64-Bins stages progressively generate more refined depth estimates. Finally, the resulting depth map is obtained, which shows a marked improvement over the initial estimate from the decoder, thereby strongly confirming the efficacy of the PQB method within the DAPM framework.


\subsection{Comparison with Monocular Depth Estimation Models on Potsdam Dataset}



\begin{table}[t]
\centering
\caption{Comparison of Depth Estimation Performance on Potsdam Dataset}

\resizebox{1\linewidth}{!}{%
\begin{tabular}{lcccc}
\toprule
\textbf{Method} & $\delta_1 \uparrow$ & $\delta_2 \uparrow$ & $\delta_3 \uparrow$ & \text{AbsRel} $\downarrow$ \\
\midrule
D3Net \cite{carvalho2018regression} & 0.601 & 0.742 & 0.830 & 0.391 \\
IM2ELEVATION \cite{zhang2019multi} & 0.638 & 0.767 & 0.839 & 0.429 \\
PLNet \cite{amirkolaee2019height} & 0.639 & 0.833 & 0.912 & 0.318 \\
BAMTL \cite{wang2020boundary} & 0.685 & 0.819 & 0.897 & 0.291 \\
InvPT \cite{ye2022inverted} & 0.673 & 0.829 & 0.904 & 0.253 \\
TaskExpert \cite{ye2023taskexpert} & 0.650 & 0.818 & 0.898 & 0.273 \\
KMNet \cite{li2025knowledge} & 0.728 & 0.863 & 0.922 & \textbf{0.242} \\
Adabins \cite{bhat2021adabins} & 0.693 & 0.844 & 0.919 & 0.324 \\
BinsFormer \cite{li2024binsformer} & 0.708 & 0.859 & 0.921 & 0.266 \\
\midrule
\textbf{DAPM(D)} & \textbf{0.732} & \textbf{0.887} & \textbf{0.944} & 0.315 \\
\bottomrule
\end{tabular}
}

\label{tab:potsdam}
\end{table}


\begin{figure}[t]
\centering
\includegraphics[width=0.95\linewidth]{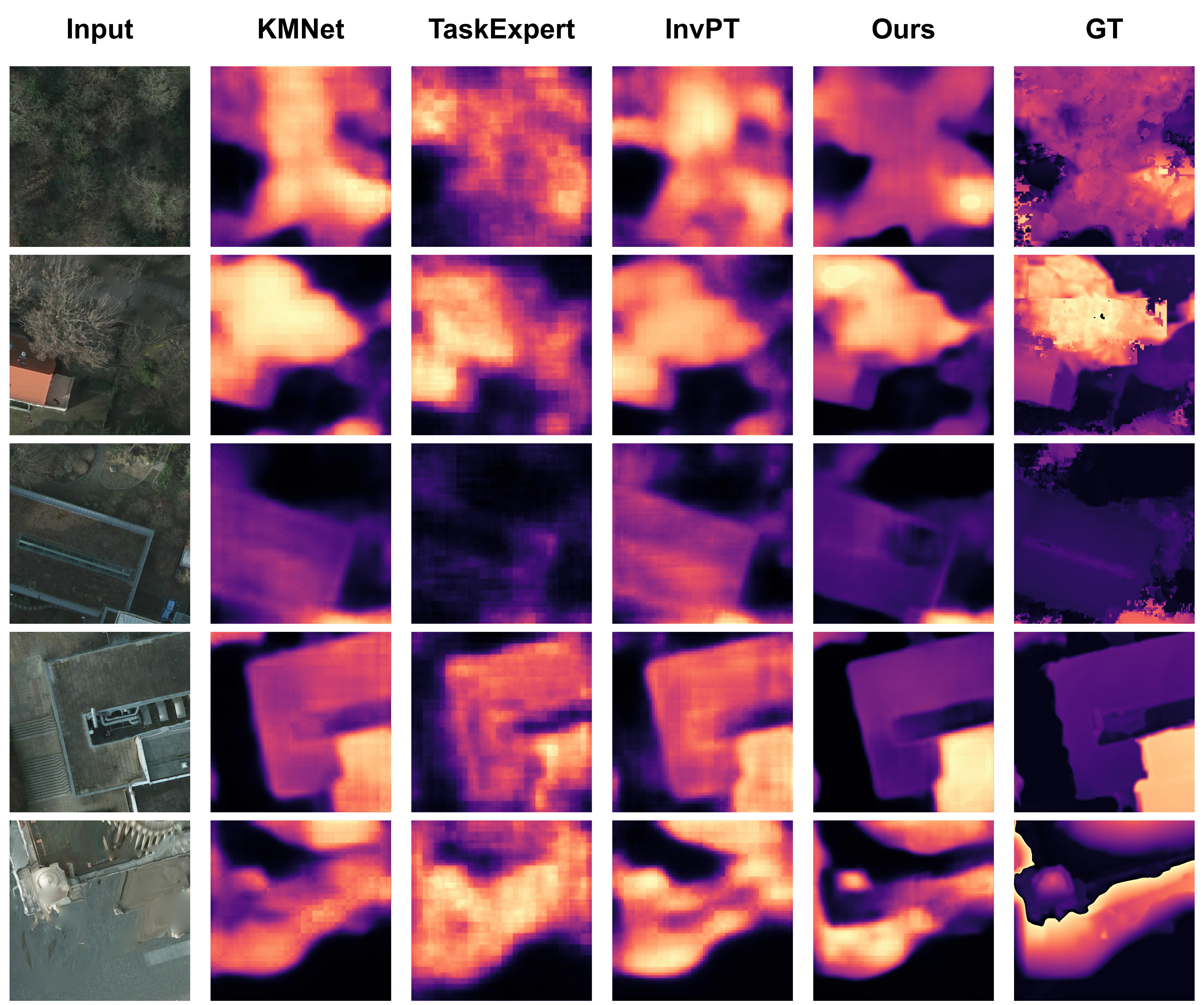}
\caption{Qualitative comparison of different depth estimation methods and the DAPM on the Potsdam dataset.}
\label{fig:potsdam_vis}
\end{figure}

To further verify the generalization capability of the proposed DAPM model beyond UAV imagery, we conducted additional experiments on the widely used Potsdam dataset, which serves as an authoritative benchmark in the remote sensing community. The Potsdam dataset contains high-resolution satellite imagery with diverse urban structures and varying illumination conditions. All images were cropped to a spatial resolution of $512\times512$ pixels and randomly split into training and validation sets with an 8:2 ratio. The experimental configuration, backbone network, and hyperparameters were kept identical to those used in the UAPD experiments to ensure a consistent evaluation protocol. Only the depth estimation task was considered during training. The quantitative results presented in Tab.~\ref{tab:potsdam} demonstrate that the proposed DAPM model achieves competitive and consistent performance across all evaluation metrics. Specifically, DAPM attains the highest accuracy scores, outperforming strong baselines such as KMNet \cite{li2025knowledge} and BinsFormer \cite{li2024binsformer} in terms of structural completeness and depth consistency. Although the AbsRel error is slightly higher than that of KMNet, DAPM maintains superior overall depth accuracy, particularly in challenging regions with complex textures or shadow occlusions. As illustrated in Fig.~\ref{fig:potsdam_vis}, the qualitative visualization further substantiates these advantages, showing that DAPM produces more coherent depth boundaries, smoother transitions in homogeneous regions, and more accurate reconstruction of fine structural details compared with other methods. 

Overall, the results on the Potsdam dataset validate the excellent cross-domain generalization of DAPM. The model consistently delivers robust performance on satellite imagery, confirming its potential applicability to diverse remote sensing depth estimation tasks.

\subsection{Comparison with Camera Pose Estimation Models}

\begin{table*}[t]
\centering
\caption{Comparison of Camera Pose Estimation Performance on UAPD Dataset}
\begin{threeparttable}
\resizebox{0.95 \linewidth}{!}{%
\begin{tabular}{l|ccc|ccc|ccc|ccc}
\toprule
\multirow{2}{*}{Method} & 
\multicolumn{3}{c|}{Roll [degrees]} & 
\multicolumn{3}{c|}{Pitch [degrees]} & 
\multicolumn{3}{c|}{FoV [degrees]} &
\multicolumn{3}{c}{Height [meters]} \\
\cmidrule(lr){2-4} \cmidrule(lr){5-7} \cmidrule(lr){8-10} \cmidrule(l){11-13}
 & error$\downarrow$ & 5°$\uparrow$ & 10°$\uparrow$ 
 & error$\downarrow$ & 5°$\uparrow$ & 10°$\uparrow$ 
 & error$\downarrow$ & 5°$\uparrow$ & 10°$\uparrow$ 
 & error$\downarrow$ & 5m$\uparrow$ & 10m$\uparrow$ \\
\midrule
DeepCalib \cite{bogdan2018deepcalib} & 5.83 & 0.23 & 0.39 & 7.06 & 0.18 & 0.34 & 7.21 & 0.19 & 0.35 & -- & -- & -- \\
Perspective Fields \cite{jin2023perspective} & 3.48 & 0.34 & 0.52 & 3.25 & 0.37 & 0.58 & 4.49 & 0.27 & 0.48 & -- & -- & -- \\
GeoCalib \cite{veicht2024geocalib} & 3.25 & 0.33 & 0.54 & 3.44 & 0.34 & 0.57 & 4.04 & 0.29 & 0.50 & -- & -- & -- \\
\midrule
DAPM(P)\tnote{*} & 3.18 & 0.34 & 0.54 & 3.25 & 0.37 & 0.67 & 3.78 & 0.31 & 0.53 & 4.44 & 0.28 & 0.50 \\
\textbf{DAPM} & \textbf{2.33} & \textbf{0.42} & \textbf{0.60} & \textbf{2.36} & \textbf{0.44} & \textbf{0.67} & \textbf{3.38} & \textbf{0.33} & \textbf{0.56} & \textbf{3.11} & \textbf{0.38} & \textbf{0.59} \\
\bottomrule
\end{tabular}
}
\begin{tablenotes}\footnotesize
\begin{minipage}{1.7\linewidth}
\item  \tnote{*}DAPM(P) indicates that the model is trained only for camera pose estimation, while DAPM is trained jointly for camera pose and depth estimation.
\end{minipage}
\end{tablenotes}

\end{threeparttable}
\label{tab:stanford_comparison}
\end{table*}

To comprehensively evaluate the effectiveness of camera pose estimation, we conducted experiments on the UAPD dataset using the unified GeoCalib \cite{veicht2024geocalib} evaluation framework. Tab.~\ref{tab:stanford_comparison} reports the quantitative results for roll, pitch, field of view (FoV), and camera height estimation. The DAPM model achieves state-of-the-art (SOTA) performance across all evaluated parameters, demonstrating its strong ability to recover accurate geometric camera poses from UAV imagery.

Unlike previous pose estimation approaches, such as DeepCalib \cite{bogdan2018deepcalib} and Perspective Fields\cite{jin2023perspective}, which primarily target ground-level or indoor scenes, DAPM is specifically designed for aerial perspectives. By incorporating hierarchical progressive supervision for camera pose learning and a depth-constrained geometric consistency loss, DAPM effectively captures the latent projection relationships between image features and camera orientation.


Furthermore, DAPM introduces an additional estimation branch for camera height, a parameter that is often neglected in existing calibration frameworks. As shown in Tab.~\ref{tab:stanford_comparison},  our pose-only baseline, DAPM(P), already demonstrates superior performance compared to leading methods like GeoCalib \cite{veicht2024geocalib} across most metrics. However, the full DAPM framework achieves further significant gains over DAPM(P) by incorporating depth-constrained geometric consistency. The joint training strategy enables DAPM to attain the lowest median errors of $2.33^{\circ}$ (Roll) and $2.36^{\circ}$ (Pitch). This corresponds to an error reduction of approximately 30\% compared to the strong competitor GeoCalib. Additionally, DAPM fills a critical gap in existing frameworks by providing reliable height estimation with a median error of only 3.11~m. 
In summary, the results demonstrate that DAPM significantly outperforms existing state-of-the-art camera calibration methods originally designed for street-view or indoor domains. The proposed model achieves high-precision estimation of roll, pitch, FoV, and height parameters on UAV imagery, highlighting its robustness and versatility in real-world aerial perception applications.

\begin{figure}[t]
\centering
\includegraphics[width=0.95\linewidth]{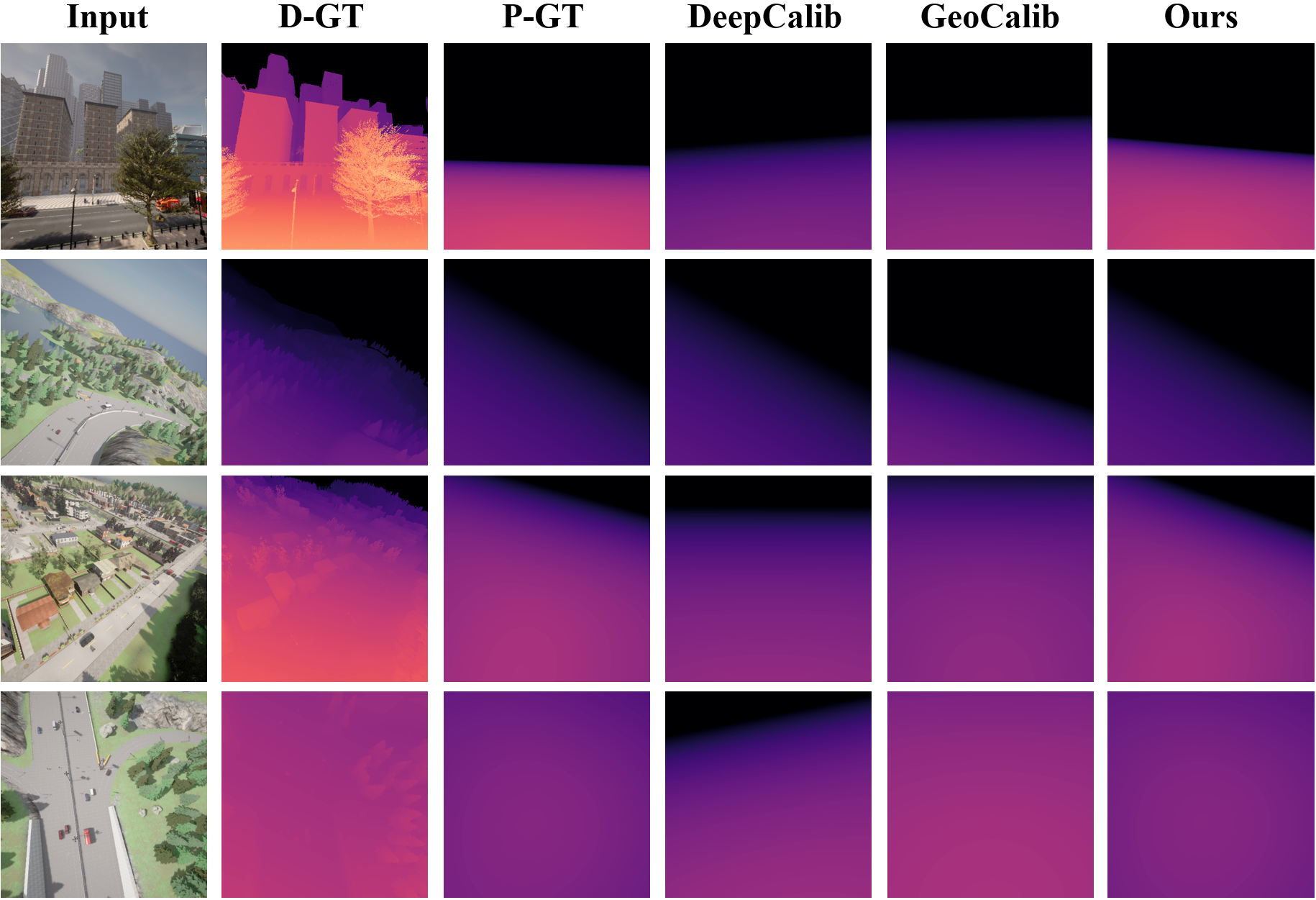}
\caption{Using the pose information obtained from different camera pose estimation methods, Ideal Ground Depth maps are generated through the IGD method for qualitative comparison. As illustrated in the figure, the pose estimation results produced by UAPM demonstrate a clearly superior performance compared to other methods.}
\label{fig:pose_vis}
\end{figure}

To enable an intuitive comparison of camera pose estimation performance, we present a visual comparison in Fig.~\ref{fig:pose_vis} using the geometric mapping method from the Ideal Ground Depth (IGD) module. It can be clearly observed that DAPM outperforms other methods in both the accuracy and stability of camera pose estimation. Furthermore, the visual comparison between the ideal ground depth and the true depth from the image reveals that the overall trend of the true depth is largely consistent with the ideal ground depth. Even in scenes with tall buildings, the ideal ground provides the model with rich prior information: rather than imposing hard constraints that could cause estimation bias, it supplies the model with camera pose information for better view representation and offers a geometric prior to distinguish the ground and sky background. This enhances the model's structural understanding of the scene and consequently improves the overall depth estimation performance.

\subsection{Ablation Experiments}

\subsubsection{Ablation Experiments for Different Modules}

\begin{table}[t]
\centering
\caption{Ablation experiment on the effectiveness of different modules, where PQB is the Progressive Quantization Bin and IGD is the Ideal Ground Depth Refinemen}
\resizebox{0.95\linewidth}{!}{%
\begin{tabular}{cc|cc|cccc}
\toprule
\multicolumn{2}{c|}{Module} & \multicolumn{2}{c|}{Depth} & \multicolumn{4}{c}{Pose} \\
\cline{1-8}
PQB & IGD & $\delta_1$ & AbsRel & Roll & Pitch & FoV & Height \\
\hline
& & 0.854 & 0.55 & 5.28 & 5.75 & 9.07 & 10.08 \\
$\checkmark$ & & 0.882 & 0.39 & 4.73 & 3.70 & 6.79 & 8.81 \\
& $\checkmark$ & 0.897 & 0.46 & 3.78 & 4.25 & 8.07 & 7.08 \\
$\checkmark$ & $\checkmark$ & \textbf{0.927} & \textbf{0.25} & \textbf{2.33} & \textbf{2.36} & \textbf{3.38} & \textbf{3.11} \\
\toprule
\end{tabular}
}

\label{tab:ae1}
\end{table}


To validate the effectiveness of each module, we conducted ablation experiments on the UAPD dataset by selectively enabling or disabling the Progressive Quantization Bins (PQB) and Ideal Ground Depth (IGD) modules. The results, summarized in Tab.~\ref{tab:ae1}, indicate that both modules positively contribute to the model’s performance in depth and camera pose estimation. Even when used individually, each module leads to noticeable improvements over the baseline, confirming its specific roles in enhancing different aspects of the predictions.

PQB enhances the model’s ability to handle depth discretization progressively, while IGD refines depth predictions and reduces pose estimation errors. The combination of both modules yields the most significant improvement, achieving the highest overall accuracy. These findings demonstrate that PQB and IGD complement each other and are essential for maximizing the model’s performance across both depth and camera pose estimation tasks.

\subsubsection{Ablation Experiments on the Contribution of Progressive Quantification Bins}

\begin{table}[htbp]
    \centering
    \caption{Ablation study on the contributions of Pose Progressive Quantification Bins(P-PQB) and Depth Progressive Quantification Bins(D-PQB) to depth and pose estimation}
    \resizebox{0.95\linewidth}{!}{%
    \begin{tabular}{cc|cc|cccc}
        \toprule
        \multicolumn{2}{c|}{PQB} & \multicolumn{2}{c|}{Depth} & \multicolumn{4}{c}{Pose} \\
        \cline{1-8}
        P-PQB & D-PQB & $\delta_1$ & AbsRel & Roll & Pitch & FoV & Height \\
        \hline
        & & 0.897 & 0.46 & 3.78 & 4.25 & 8.07 & 7.08 \\
        $\checkmark$ & & 0.893 & 0.41 & 2.97 & 3.09 & 3.83 & 3.28 \\
         & $\checkmark$ & 0.916 & 0.28 & 3.45 & 4.37 & 5.62 & 4.73 \\
        $\checkmark$ & $\checkmark$ & \textbf{0.927} & \textbf{0.25} & \textbf{2.33} & \textbf{2.36} & \textbf{3.38} & \textbf{3.11} \\
        \toprule
    \end{tabular}
    }
    \label{tab:pqbae}
\end{table}

To further disentangle the roles of the Progressive Quantization Bins applied to pose progressive quantification bins(P-PQB) and depth progressive quantification bins(D-PQB), we conducted ablation experiments with different combinations of these modules, as summarized in Tab.~\ref{tab:pqbae}. The results show that enabling P-PQB alone yields clear gains in both depth and pose estimation, indicating that quantized geometric cues from pose prediction can enhance the spatial reasoning required for depth regression. Conversely, using D-PQB alone substantially improves depth estimation accuracy but offers limited benefits to pose estimation, suggesting that quantized depth priors are less directly transferable to camera pose recovery. When both P-PQB and D-PQB are jointly employed, the model achieves the best performance across all metrics for both tasks. This demonstrates that the two quantization pathways provide complementary supervisory signals: P-PQB strengthens geometric consistency, while D-PQB refines depth discretization. Their synergy further stabilizes optimization and enriches multi-scale geometric representations.

\subsubsection{Ablation Experiments on the Effectiveness of IGD}

\begin{table}[htbp]
    \centering
    \caption{Ablation study on the effectiveness of the feature refinement and loss constraint in IGD}
    \resizebox{0.95\linewidth}{!}{%
    \begin{tabular}{cc|cc|cccc}
        \toprule
        \multicolumn{2}{c|}{IGD} & \multicolumn{2}{c|}{Depth} & \multicolumn{4}{c}{Pose} \\
        \cline{1-8}
        Feature & Loss & $\delta_1$ & AbsRel & Roll & Pitch & FoV & Height \\
        \hline
        & & 0.882 & 0.39 & 4.73 & 3.70 & 6.79 & 8.81 \\
        $\checkmark$ & & 0.914 & 0.29 & 3.11 & 4.58 & 3.89 & 4.59 \\
         & $\checkmark$ & 0.908 & 0.33 & 2.98 & 2.94 & \textbf{3.31} & 3.89 \\
        $\checkmark$ & $\checkmark$ & \textbf{0.927} & \textbf{0.25} & \textbf{2.33} & \textbf{2.36} & 3.38 & \textbf{3.11} \\
        \toprule
    \end{tabular}
    }
    \label{tab:IGDae}
\end{table}

We also investigate the impact of the Ideal Ground Depth (IGD) module by isolating its two major components, the feature-level refinement and the loss-level constraint, as shown in Tab.~\ref{tab:IGDae}. Introducing feature-level IGD alone improves depth accuracy and yields moderate gains in roll and FoV estimation, indicating that refined feature representations help the network better capture structural depth cues. Meanwhile, applying the IGD loss alone results in marked improvements in pose parameters, especially pitch and roll, demonstrating the effectiveness of the Ideal Ground Depth. The full IGD configuration, combining both feature refinement and loss constraints, consistently achieves the best results across all depth and pose metrics. This validates that the two components reinforce each other: feature refinement enhances the quality of intermediate depth representations, while the loss constraint stabilizes optimization by guiding predictions toward physically plausible geometric structures. Their integration thus provides a more comprehensive refinement mechanism, ultimately strengthening both depth and pose estimation performance.

\subsubsection{Ablation Experiments for Different Quantization Levels in Depth Estimation}

To investigate the influence of the quantization design in the Progressive Quantization Bins (PQB) on depth estimation, we conducted ablation experiments with varying numbers of layers and bins, as summarized in Tab.~\ref{tab:component_analysis}. The results show that introducing quantization layers notably enhances the accuracy of depth estimation compared with the baseline model without quantization. Progressive quantization with gradually increasing bin numbers further improves performance, confirming that hierarchical quantization helps the network learn more discriminative depth representations. However, overly fine-grained quantization does not necessarily yield additional gains, suggesting that excessive bin resolution may introduce redundant or unstable information. These findings demonstrate that an appropriately progressive quantization structure achieves the best balance between representation precision and generalization in depth estimation.

\begin{table}[htbp]
    \centering
    \caption{Ablation experiment on the effect of different numbers of bins in the quantification layer on depth estimation}
    \begin{tabular}{ccc|cc}
        \toprule
        \multicolumn{3}{c|}{Number of Bins} & \multicolumn{2}{c}{Depth} \\
        \cline{1-5}
        Layer 1 & Layer 2 & Layer 3 & $\delta_1$ & AbsRel \\
        \hline
        -- & -- & -- & 0.897 & 0.46 \\
        2 & -- & -- & 0.899 & 0.34 \\
        2 & 16 & -- & 0.902 & 0.28 \\
        2 & 16 & 16 & 0.920 & 0.27 \\
        2 & 64 & 64 & 0.922 & 0.32 \\
        2 & 16 & 64 & \textbf{0.927} & \textbf{0.25} \\
        \toprule
    \end{tabular}
    \label{tab:component_analysis}
\end{table}

\subsubsection{Ablation Experiments for Different Quantization Levels in Camera Pose Estimation}

We further evaluate the effect of quantization in PQB on camera pose estimation, as reported in Tab.~\ref{tab:quant_analysis}. Similar to the depth results, incorporating quantization layers significantly reduces pose estimation errors across all parameters, including roll, pitch, field of view (FoV), and height. The use of multiple quantization layers with progressively increasing bin numbers leads to more stable and accurate pose recovery, indicating that the hierarchical quantization strategy enhances the representation of spatial geometry and camera motion. In contrast, configurations with excessively large bin counts bring marginal benefits or even slight degradation, implying that over-quantization may reduce robustness. Overall, these results confirm the effectiveness of progressive quantization in improving both the precision and stability of pose estimation within UAV-based perception tasks.

\begin{table}[htbp]
    \centering
    \caption{Ablation experiment on the effect of different numbers of bins in the quantification layer on camera pose estimatio}
    \begin{tabular}{cc|cccc}
        \toprule
        \multicolumn{2}{c|}{Number of Bins} & \multicolumn{4}{c}{Pose} \\
        \cline{1-6}
        Layer 1 & Layer 2 & Roll & Pitch & FoV & Height \\
        \hline
        -- & -- & 3.78 & 4.25 & 8.07 & 7.08 \\
        16 & -- & 3.52 & 3.85 & 4.07 & 4.33 \\
        16 & 16 & 2.66 & 3.16 & \textbf{3.35} & 3.58 \\
        64 & 64 & 2.61 & 3.26 & 3.82 & \textbf{2.93} \\
        16 & 64 & \textbf{2.33} & \textbf{2.36} & 3.38 & 3.11 \\
        \toprule
    \end{tabular}
    \label{tab:quant_analysis}
\end{table}

\section{Conclusion}

In this paper, we presented DAPM, a novel framework that pioneers arbitrary-viewpoint depth estimation for UAV imagery by leveraging the geometric relationship between viewing angles and depth distributions to concurrently optimize camera pose. Validated on our newly constructed UAPD dataset containing 42k images with continuous pose variations, DAPM achieves state-of-the-art performance, demonstrating substantial accuracy improvements in both depth and pose estimation tasks. This study not only fills the research gap in pose-aware aerial depth estimation but also establishes a foundation for future work, which we plan to extend towards real-world data collection, complex terrain adaptability, and broader UAV vision applications.




\section*{Acknowledgments}
This work was supported by the National Natural Science Foundation of China under Grant 62425115.


\bibliographystyle{ieeetr}
\bibliography{refs}

\ifCLASSOPTIONcaptionsoff
\newpage
\fi

\end{document}